\pdfoutput=1

\documentclass{article}

\PassOptionsToPackage{numbers, sort&compress}{natbib}

\usepackage[utf8]{inputenc} 
\usepackage[T1]{fontenc}    
\usepackage{hyperref}       
\usepackage{url}            
\usepackage{booktabs}       
\usepackage{amsfonts}       
\usepackage{nicefrac}       
\usepackage{microtype}      
\usepackage{color}
\usepackage[dvipsnames]{xcolor}
\usepackage{graphicx}
\usepackage{floatrow}
\usepackage{subfig}
\usepackage[percent]{overpic}
\usepackage{amsmath}
\usepackage{amssymb}
\usepackage{enumitem}
\usepackage{ulem}
\usepackage{comment}
\usepackage{cleveref}

\usepackage[accepted]{icml2020}

\DeclareFloatVCode{afterfloat}{}
\newlength{\defaultlength}
\newlength{\adjlength}
\icmltitlerunning{On the relationship between class selectivity, dimensionality, and robustness}

\newif\ifshow
\showtrue
\showfalse

\ifshow
\newcommand{\ari}[1]{\textcolor{blue}{[\textbf{Ari:} #1]}}
\newcommand{\mat}[1]{\textcolor{Aquamarine}{[\textbf{Mat:} #1]}}
\newcommand{\todo}[1]{\textcolor{red}{[TO DO: #1]}}
\else
\newcommand{\ari}[1]{}
\newcommand{\mat}[1]{}
\newcommand{\todo}[1]{}
\fi

\newcommand*{\fullref}[1]{\hyperref[{#1}]{\autoref*{#1} \nameref*{#1}}}

\begin{document}

\twocolumn[
\icmltitle{On the relationship between class selectivity, dimensionality, and robustness}



\icmlsetsymbol{equal}{*}

\begin{icmlauthorlist}
\icmlauthor{Matthew L. Leavitt}{fb,res} 
\icmlauthor{Ari S. Morcos}{fb}
\end{icmlauthorlist}

\icmlaffiliation{fb}{Facebook AI Research, Menlo Park, California}
\icmlaffiliation{res}{Work performed as part of the Facebook AI Residency program}

\icmlcorrespondingauthor{Matthew L. Leavitt}{ito@fb.com}
\icmlcorrespondingauthor{Ari S. Morcos}{arimorcos@fb.com}


\vskip 0.3in
]



\printAffiliationsAndNotice{}  

\begin{abstract}

While the relative trade-offs between sparse and distributed representations in deep neural networks (DNNs) are well-studied, less is known about how these trade-offs apply to representations of semantically-meaningful information. Class selectivity—the variability of a unit’s responses across data classes or dimensions—is one way of quantifying the sparsity of semantic representations. Given recent evidence showing that class selectivity can impair generalization, we sought to investigate whether it also confers robustness (or vulnerability) to perturbations of input data. We found that mean class selectivity predicts vulnerability to naturalistic corruptions; networks regularized to have lower levels of class selectivity are more robust to corruption, while networks with higher class selectivity are more vulnerable to corruption, as measured using Tiny ImageNetC and CIFAR10C. In contrast, we found that class selectivity increases robustness to multiple types of gradient-based adversarial attacks. To examine this difference, we studied the dimensionality of the change in the representation due to perturbation, finding that decreasing class selectivity increases the dimensionality of this change for both corruption types, but with a notably larger increase for adversarial attacks. These results demonstrate the causal relationship between selectivity and robustness and provide new insights into the mechanisms of this relationship.

\end{abstract}
\section{Introduction} \label{sec:introduction}

Methods for understanding deep neural networks (DNNs) often attempt to find individual neurons or small sets of neurons that are representative of a network's decision \citep{erhan_visualizing_2009, zeiler_deconvolution_2014, karpathy_visualizing_rnns_2016, amjad_understanding_2018, lillian_ablation_2018, dhamdhere_conductance_2019, olah_zoom_2020}. However, recent work has shown that these neurons can be irrelevant, or even detrimental to network performance, emphasizing the importance of examining distributed representations for understanding DNNs \citep{morcos_single_directions_2018, donnelly_interpretability_2019,dalvi_neurox_2019, leavitt_selectivity_2020}. In parallel, work on robustness seeks to build models that are robust to perturbed inputs \citep{szegedy_intriguing_2013,carlini_adversarial_2017,carlini_towards_2017,vasiljevic_examining_2016,kurakin_adversarial_2017,gilmer_motivating_2018,zheng_improving_2016}. In this work we pursue a series of experiments investigating the causal role of class selectivity\footnote{Class selectivity is typically defined as how different a neuron’s responses are across different classes of stimuli or data samples. Another way to conceptualize class selectivity is as a measure of the sparsity of a network's semantic representations.} on adversarial and corruption robustness in DNNs\footnote{As articulated by \citet{hendrycks_tinyimagenetc_2019}, corruption robustness measures a classifier's performance on low-quality or naturalistically-perturbed inputs, and thus is an "average-case" measure. Adversarial robustness measures a classifier's performance on small, additive perturbations that are tailored to the classifier, and thus is a "worst-case" measure. \ari{space-permitting this shouldn't be a footnote}}. To do so, we used a recently-developed class selectivity regularizer \citep{leavitt_selectivity_2020} to directly modify the amount of class selectivity learned by DNNs, and examined how this affected the DNNs' vulnerability to corruption and adversarial perturbations. Our findings are as follows:

\vspace{-3mm}
\newcommand{\contspace}{\vspace{-1mm}}
\begin{itemize}[leftmargin=1.2em]
\setlength{\labelsep}{\defaultlength}
    \contspace
    \item Decreasing class selectivity decreases vulnerability to corruption in both ResNet18 trained on Tiny Imagenet and ResNet20 trained on CIFAR10 across nearly all tested corruptions. 
    \vspace{-2mm}
    \item In contrast to its impact on naturalistic corruptions, decreasing class selectivity \textit{increases} vulnerability to gradient-based adversarial attacks in both tested models. 
    \vspace{-5mm}
    \item The dimensionality of activation changes caused by corruption markedly increases in early layers for both perturbation types, but is larger for adversarial attacks and low-selectivity networks. This implies that high-dimensional representations may induce adversarial vulnerability.
    \vspace{-3mm}
\end{itemize}
\setlength{\labelsep}{\adjlength}


Our results demonstrate that changes in the sparsity of semantic representations can confer robustness to naturalistic or adversarial perturbations, but not both simultaneously. They also highlight the roles of class selectivity and representational dimensionality in mediating a trade-off between worst-case and average-case perturbation robustness. More generally, our results encourage further investigation of the stability and robustness of sparse vs. distributed representations, and stress the importance of verifying that improvements in robustness are not zero-sum.










%

\section{Approach} \label{sec:approach}


A detailed description is provided in Appendix \ref{appendix:models_training_etc}.
\vspace{-3mm}
\paragraph{Models and training protocols}\label{sec:approach_models_datasets}
Our experiments were performed on ResNet18 \citep{he_resnet_2016} trained on Tiny ImageNet \citep{tiny_imagenet}, and ResNet20 \citep{he_resnet_2016} trained on CIFAR10 \citep{krizhevsky_cifar10_2009}.
\vspace{-3mm}
\paragraph{Datasets for naturalistic corruptions}\label{sec:approach_cifar10c_tinyimagenetc}
To evaluate robustness to naturalistic corruptions, we tested our networks on CIFAR10C and Tiny ImageNetC, two benchmark datasets created for this purpose \citep{hendrycks_tinyimagenetc_2019}. We average across all corruption types and severities (see Appendix \ref{sec:apx_data} for details) when reporting corrupted test accuracy.
\vspace{-3mm}
\paragraph{Class selectivity regularization}\label{sec:approach_selectivity_regularizer}
We quantify class selectivity in individual units using \citep{morcos_single_directions_2018}'s method. At every ReLU, the activation in response to a single sample was averaged across all elements of the filter map (which we refer to as a "unit"). The class-conditional mean activation was then calculated across data samples. We then computed:
\vspace{-2mm}
\begin{equation}
    selectivity = \frac{\mu_{max} - \mu_{-max}}{\mu_{max} + \mu_{-max}}
\end{equation}
where $\mu_{max}$ is the largest class-conditional mean activation and $\mu_{-max}$ is the mean response to the remaining classes. The selectivity index ranges from 0 to 1. A unit with identical average activity for all classes would have a selectivity of 0, and a unit that only responds to a single class would have a selectivity of 1.

We used \citep{leavitt_selectivity_2020}'s class selectivity regularizer to control the levels of class selectivity learned by units in a network during training. The regularizer, $-\alpha\mu_{SI}$, is added to the cross-entropy loss, and comprises two terms: $\mu_{SI}$, the network selectivity, and $\alpha$, the regularization scale. Network selectivity is obtained by computing the mean selectivity index across units in each layer, then computing the mean across layers. Negative values of the regularization scale, $\alpha$, discourage class selectivity in individual units, while positive values promote it. The magnitude of $\alpha$ controls the contribution of the network selectivity to the overall loss. During training, class selectivity was computed for each minibatch. For the analyses presented here, class selectivity was computed across the entire clean test set.
\vspace{-3mm}
\paragraph{Adversarial testing}\label{sec:approach_adversarial}
We tested our models' adversarial robustness using two methods. The fast gradient sign method (FGSM) \citep{goodfellow_explaining_2015} is a simple attack that computes the gradient of the loss with respect to the input image, then scales the image's pixels (within some bound) in the direction that increases the loss. The second method, projected gradient descent (PGD) \citep{kurakin_adversarial_2016, madry_towards_2018}, is an iterated version of FGSM.
\vspace{-3mm}
\paragraph{Quantifying dimensionality}\label{sec:approach_dimensionality}
Estimates of dimensionality were obtained for each layer in a network by applying PCA to the layer's activation matrix for the clean test data, counting the number of dimensions necessary to explain 90\% of the variance, then dividing by the total number of dimensions (i.e. the fraction of total dimensionality). The same procedure was applied to compute the dimensionality of perturbation-induced changes in representations, except the activations for a perturbed data set were subtracted from the corresponding clean activations prior to applying PCA.
\begin{figure*}[!thp]
    \centering
    \begin{subfloatrow*}
        \sidesubfloat[]{
        \label{fig:acc_abs_mean_tinyimagentc}
            \includegraphics[width=0.27\textwidth]{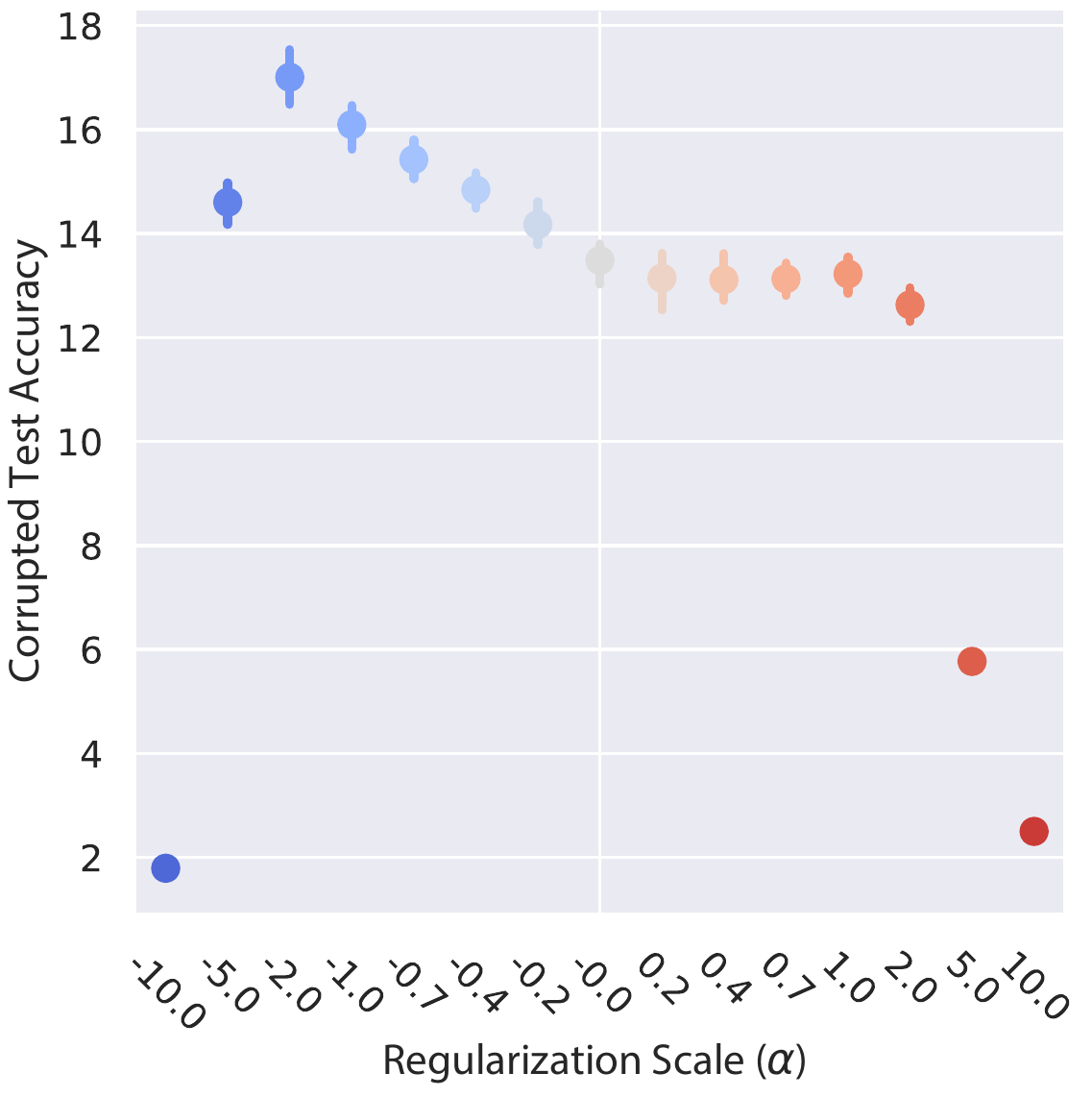}
        }
        \hspace{-1mm}
        \sidesubfloat[]{
            \label{fig:acc_rel_mean_tinyimagentc}
            \includegraphics[width=0.295\textwidth]{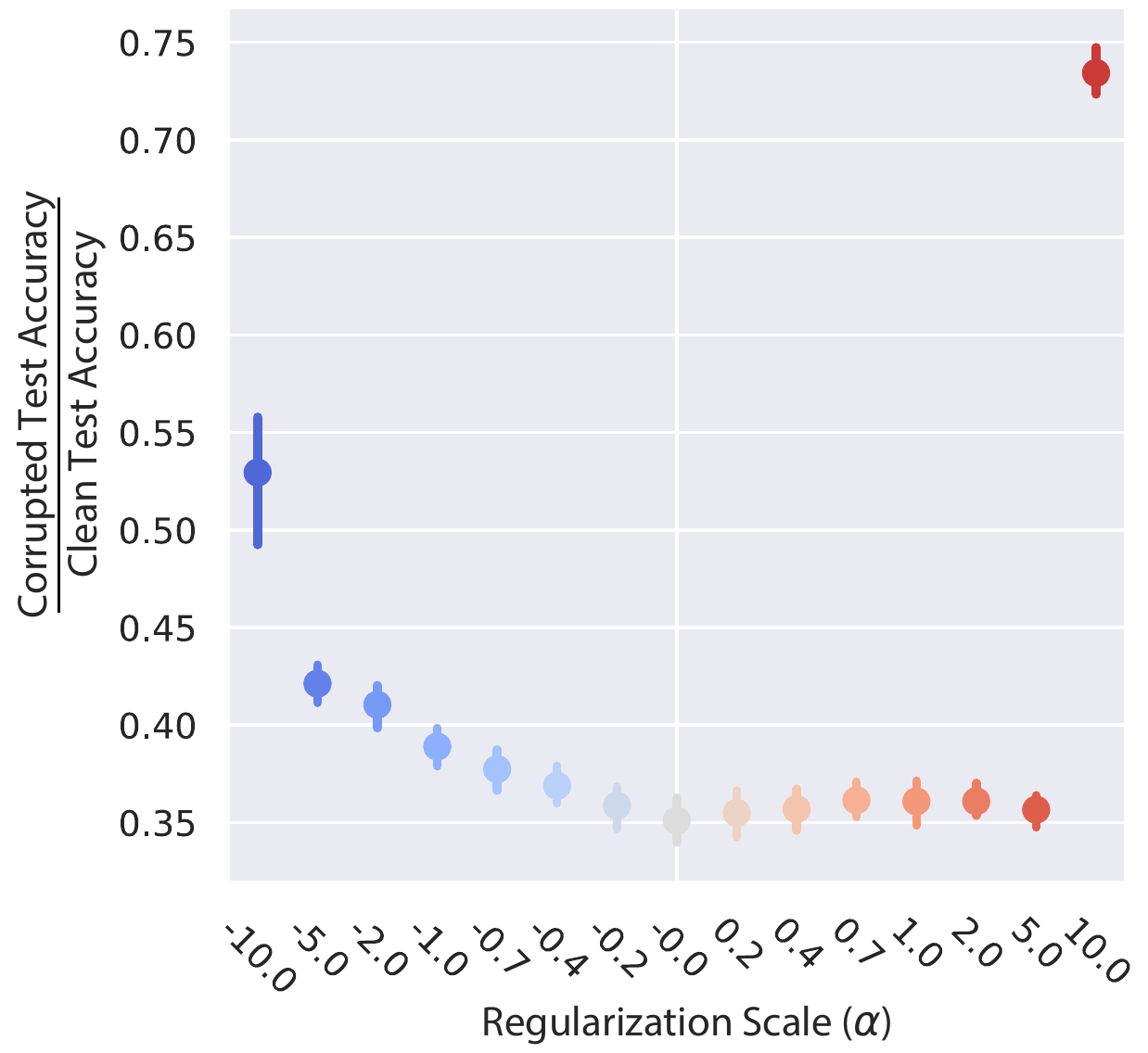}
        }
        \hspace{3mm}
        \sidesubfloat[]{
            \label{fig:perturbation_types_tinyimagenetc}
            \includegraphics[width=0.3\textwidth]{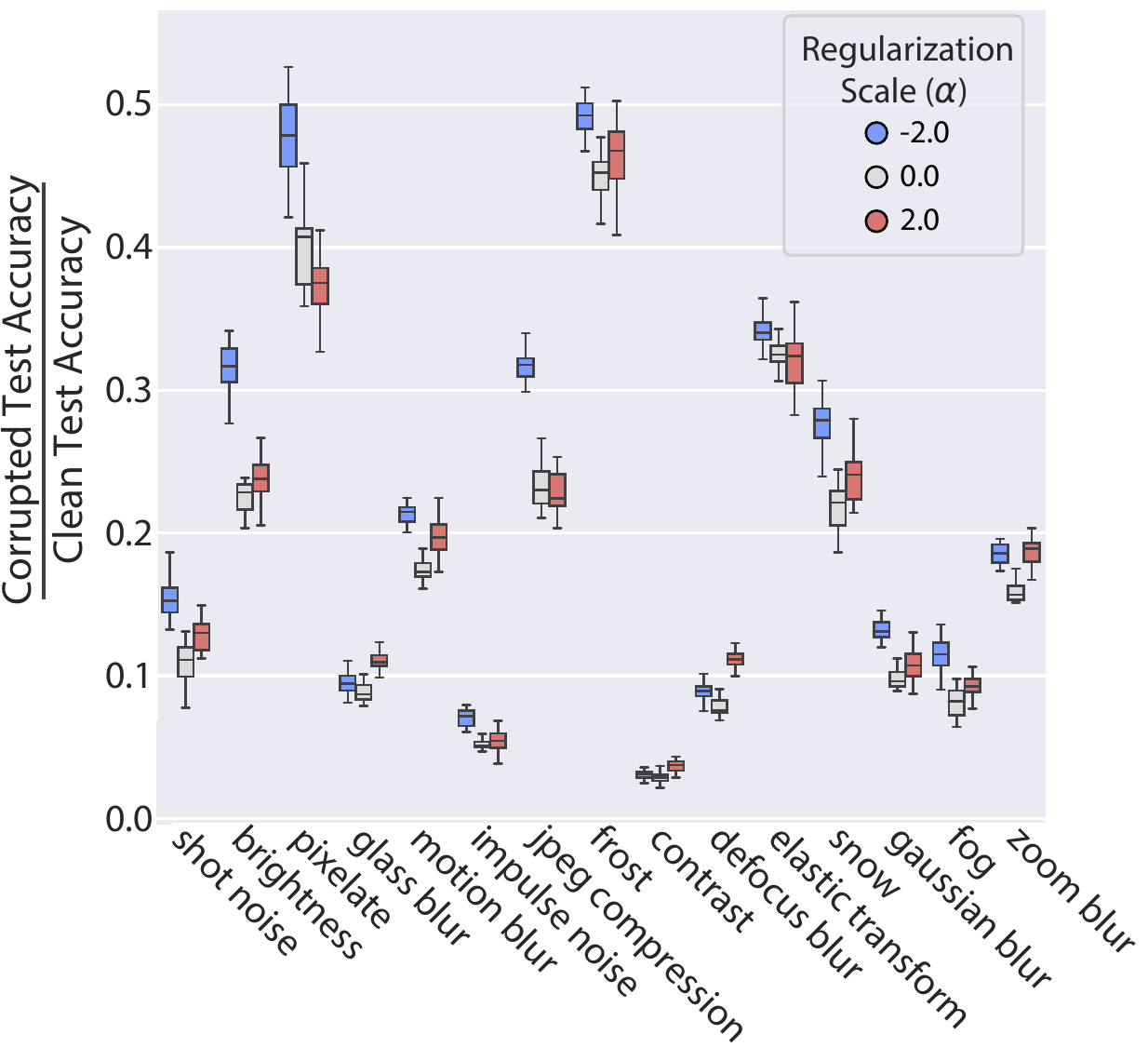}
        }
    \end{subfloatrow*}
    \vspace{-8mm}
    \caption{\textbf{Reducing class selectivity confers corruption robustness.} (\textbf{a}) Mean test accuracy across all corruptions and severities (y-axis) as a function of class selectivity regularization scale ($\alpha$; x-axis). Negative $\alpha$ lowers selectivity, positive $\alpha$ increases selectivity, and the magnitude of $\alpha$ changes the strength of the effect (see Figure \ref{fig:si_sel_reg} and Appendix \ref{sec:apx_class_selectivity}). (\textbf{b}) Corrupted test accuracy normalized by clean test accuracy (y-axis) as a function of $\alpha$ (x-axis). (\textbf{c}) Normalized test accuracy (y-axis) for all 15 Tiny ImageNetC corruption types at severity = 5 (x-axis) for three example values of $\alpha$. Results shown are for ResNet18 trained on Tiny ImageNet, tested on Tiny ImageNetC. See Figure \ref{fig:si_acc_cifar10c} for CIFAR10C results.}
    \label{fig:acc_tinyimagenetc}
    \vspace{2mm}    
\end{figure*}

\begin{figure*}[!t]
    \centering
    \begin{subfloatrow*}
        \sidesubfloat[]{
        \label{fig:fgsm_resnet18}
            \includegraphics[width=0.3\textwidth]{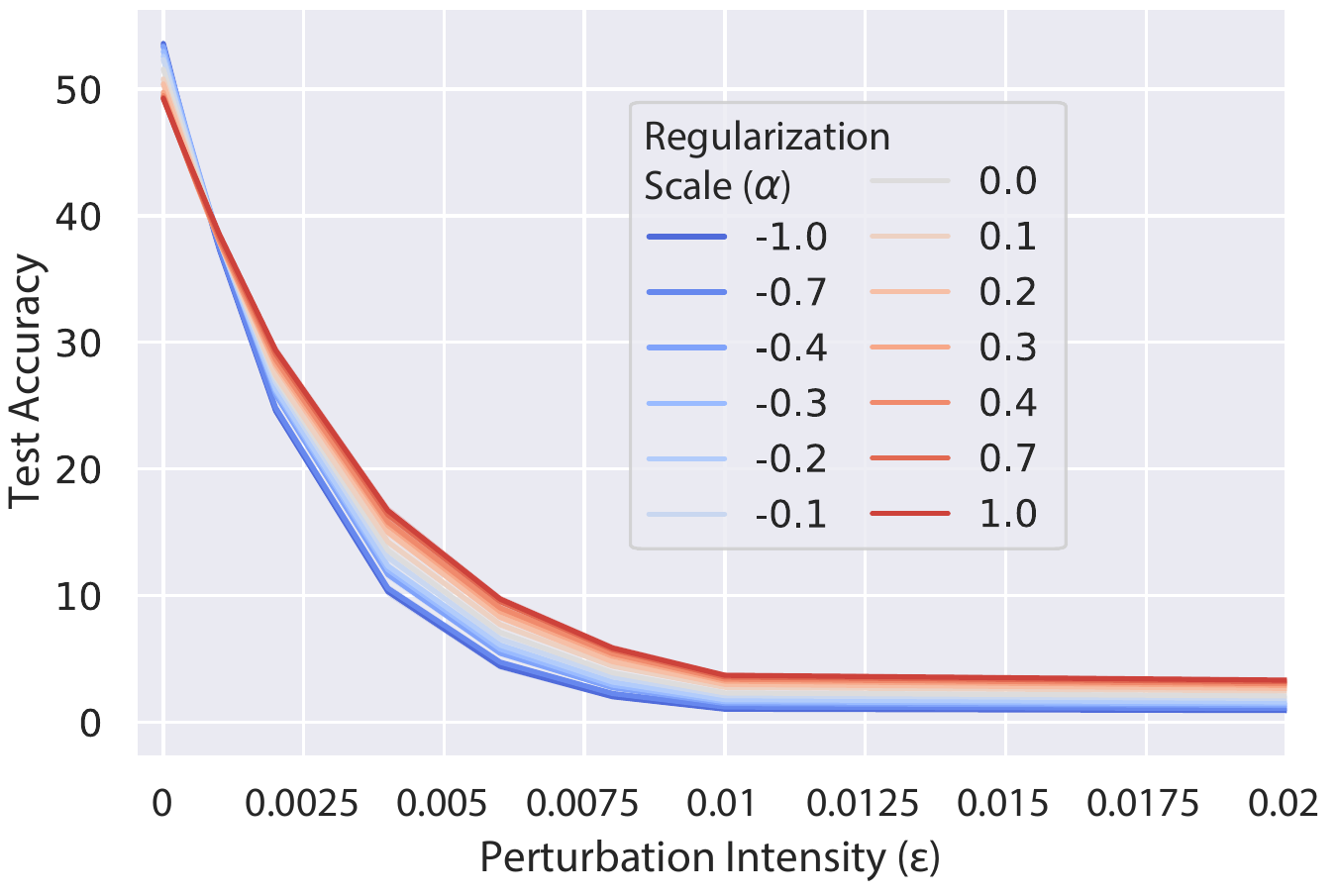}
        }
        \hspace{-4mm}
        \sidesubfloat[]{
            \label{fig:pgd_resnet18}
            \includegraphics[width=0.3\textwidth]{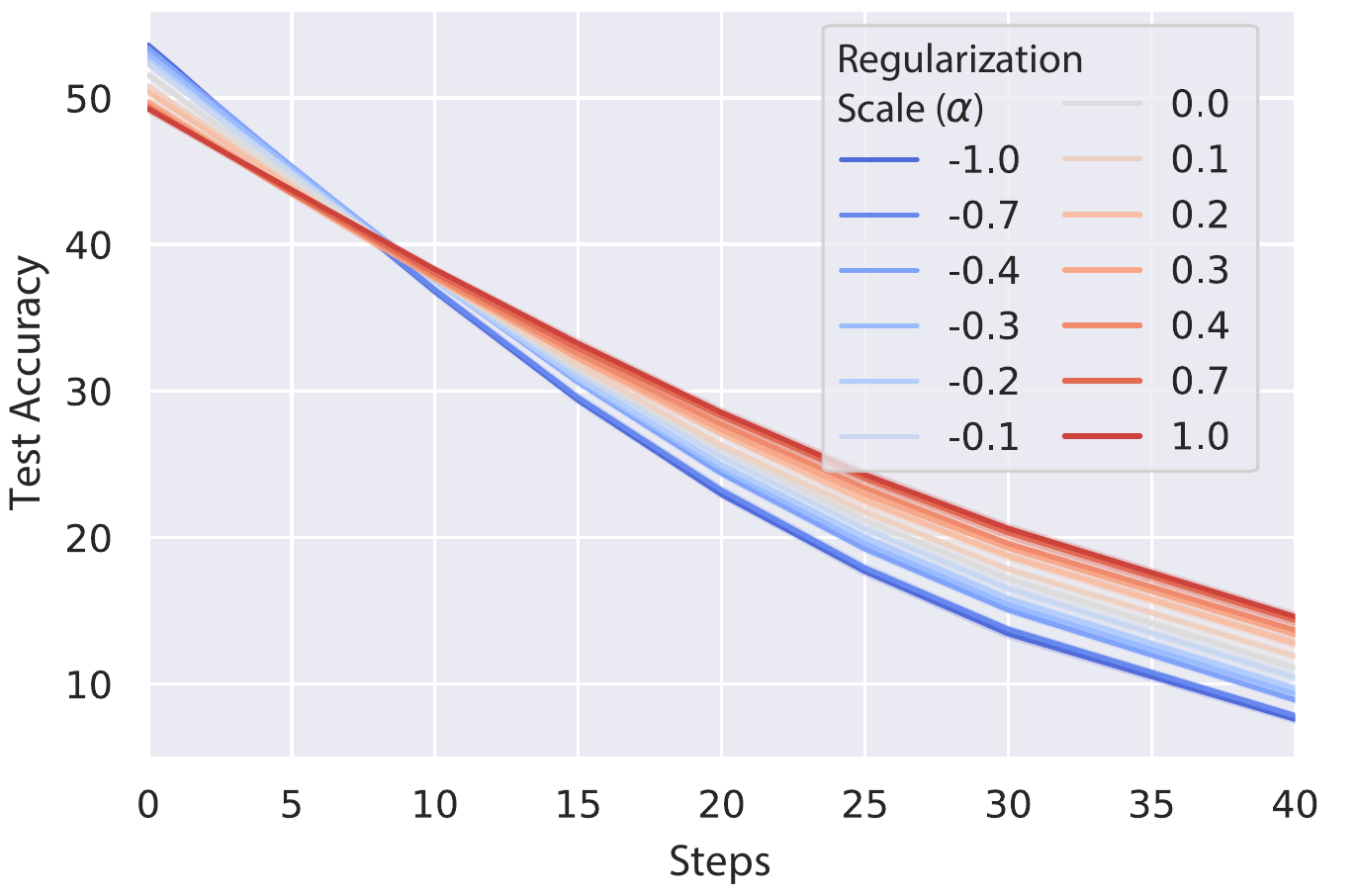}
        }
        \hspace{3mm}
        \sidesubfloat[]{
            \label{fig:jacobian_resnet18}
            \includegraphics[width=0.29\textwidth]{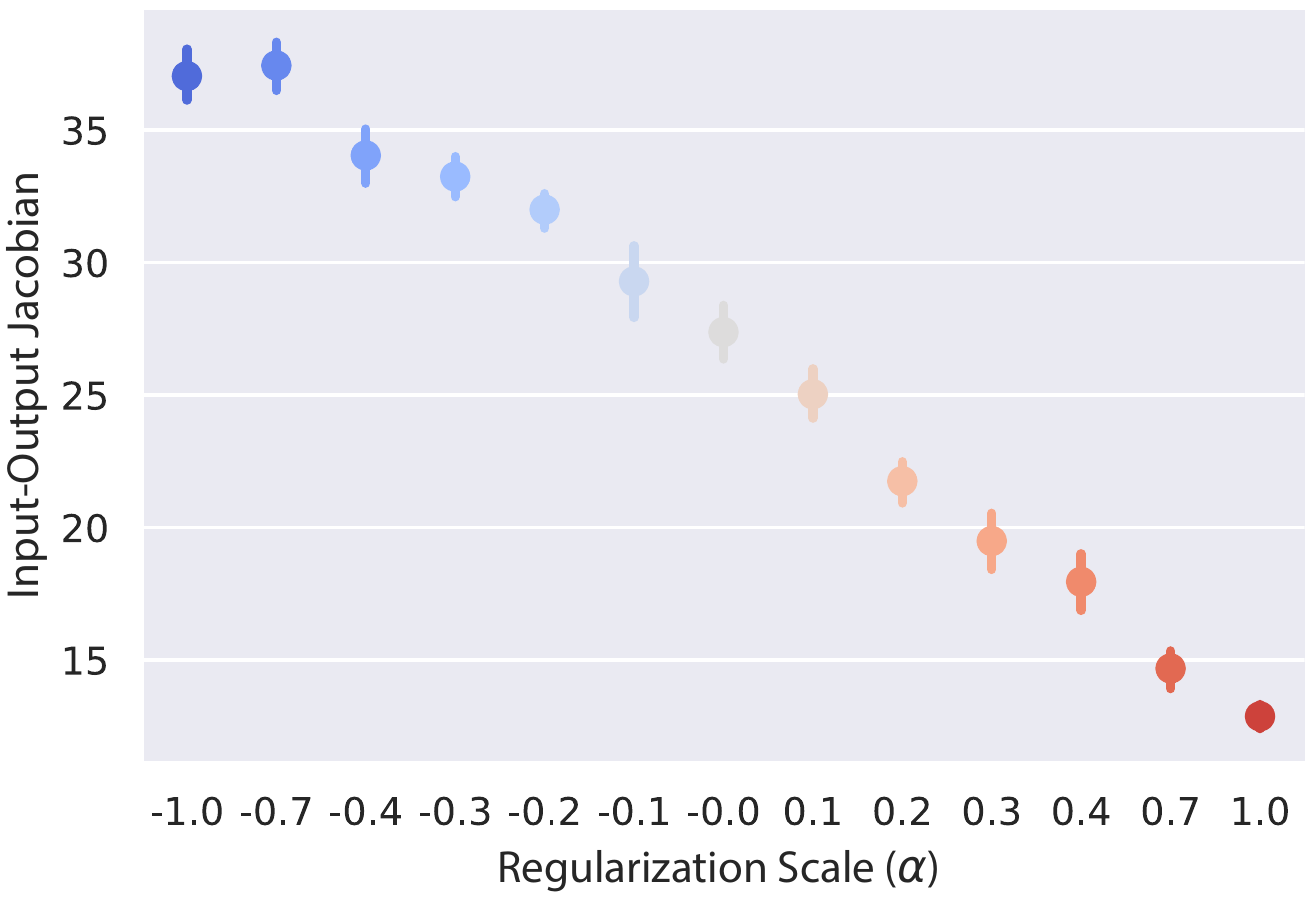}
        }
    \end{subfloatrow*}
    \vspace{-6mm}
    \caption{\textbf{Reducing class selectivity increases adversarial vulnerability.} (\textbf{a}) Test accuracy (y-axis) as a function of perturbation intensity ($\epsilon$; x-axis) and class selectivity regularization scale ($\alpha$; color) for the FGSM attack. (\textbf{b}) Test accuracy (y-axis) as a function of adversarial optimization steps (x-axis) and $\alpha$ for the PGD attack. (\textbf{c}) Network stability, as measured with input-output Jacobian (y-axis) as a function of $\alpha$. All results are for ResNet18 trained on Tiny ImageNet. See Figure \ref{fig:si_adv_cifar10} for ResNet20 results.}
    \label{fig:adv_tinyimagenet}
\end{figure*}

\section{Results} \label{sec:results}
\subsection{Class selectivity causes vulnerability to naturalistic corruptions} \label{sec:results_corruption_robustness}

To understand the relationship between semantic sparsity and robustness, we first focused on its impact on naturalistic corruptions of the input data. 
We used a recently-introduced method (\citep{leavitt_selectivity_2020}; Appendix \ref{sec:apx_class_selectivity}) to modulate the amount of class selectivity—a measure of the sparsity with which semantic information is represented—learned by DNNs\footnote{See Figure \ref{fig:si_sel_reg} for the effects of selectivity regularization.}. We then examined how this affected performance on Tiny ImageNetC and CIFAR10C, two benchmark datasets for naturalistic corruptions (Approach \ref{sec:approach}).

Changing the level of class selectivity across neurons in a network could have one of three possible effects on corruption robustness: If concentrating semantic representations into fewer neurons provides fewer potent dimensions on which corrupted inputs can act, then increasing class selectivity should confer networks with robustness to corruptions, while reducing class selectivity should render networks more vulnerable. 
Alternatively, if distributing semantic representations across more units dilutes the changes induced by corrupted inputs, then reducing class selectivity should reduce a network's vulnerability to corruptions, while increasing class selectivity should increase corruption vulnerability. 
Finally, if class selectivity regularization changes representations along dimensions that are orthogonal to those upon which corruptions act, we should fail to see any relationship between a network's class selectivity and its robustness to naturalistic corruptions.

We found that class selectivity leads to increased vulnerability to naturalistic corruptions for both Tiny ImageNetC (Figure \ref{fig:acc_tinyimagenetc}) and CIFAR10C (Figure \ref{fig:si_acc_cifar10c}). For example, decreasing class selectivity increases the mean test accuracy on corrupted inputs by as much as 4\% on Tiny ImageNetC (Figure \ref{fig:acc_abs_mean_tinyimagentc}). To control for the possibility that this improvement was simply due to changes in clean test accuracy, we also analyzed the ratio of corrupted to clean test accuracy, finding qualitatively similar results (Figures \ref{fig:acc_rel_mean_tinyimagentc} and \ref{fig:si_acc_rel_mean_cifar10c}). This effect was largely consistent across corruption types\footnote{Regularizing against selectivity improves corruption robustness in 12 of 15 corruption types in Tiny ImageNetC for ResNet18 (Figure \ref{fig:perturbation_types_tinyimagenetc}) and in 15 of 19 corruption types in CIFAR10C for ResNet20 (Figure \ref{fig:si_perturbation_types_cifar10c}) when measured using normalized test accuracy, and in all corruption types in both models when measured using absolute test accuracy (Figures \ref{fig:si_accuracy_every_corruption_tinyimagenetc} and \ref{fig:si_accuracy_every_corruption_cifar10c}).}. Together these results demonstrate that reduced class selectivity confers robustness to naturalistic corruptions, implying that distributing semantic representations across neurons dilutes the changes induced by corrupted inputs.

\subsection{Class selectivity imparts adversarial robustness}
\label{sec:results_adversarial_vulnerability}

We showed that the sparsity of a network's semantic representations, as measured with class selectivity, is causally related to a network's robustness to naturalistic corruptions. But naturalistic corruptions are an average-case measure of a DNN's perturbation-robustness, so a complementary question remains unanswered: how does the sparsity of semantic representations affect \textit{worst-case} robustness? We addressed this question by testing our class selectivity-regularized networks on inputs that had been adversarially-perturbed using using one of two white-box methods (see Approach \ref{sec:approach_adversarial}).


Surprisingly and unlike naturalistic corruptions, decreasing class selectivity \textit{increases} vulnerability to adversarial perturbation for both Tiny ImageNet and CIFAR10 (Figures \ref{fig:adv_tinyimagenet} and \ref{fig:si_adv_cifar10}). 
This result demonstrates that sparse semantic representations are less vulnerable to adversarial perturbation than distributed semantic representations.

To understand \textit{why} sparse semantic representations are more robust to adversarial corruptions, we analyzed each network's input-output Jacobian \citep{novak_sensitivity_2018,sokolic_robust_2017,rifai_contractive_2011,sho_jacobian_2019}, which is  
proportional to its stability—a large-magnitude Jacobian means that a small change to the network's input will cause a large change to its output.
If class selectivity induces adversarial robustness by increasing network stability, then networks with lower class selectivity should have smaller Jacobians. But if increased class selectivity induces adversarial robustness through alternative mechanisms, then class selectivity should have no effect on the Jacobian. 

We found that the magnitude of the input-output Jacobian is inversely proportional to class selectivity for both ResNet18 (Figure \ref{fig:jacobian_resnet18}) and ResNet20 (Figure \ref{fig:si_jacobian_resnet20}), indicating that distributed semantic representations are more vulnerable to adversarial perturbation because they are less stable than sparse semantic representations.


\begin{figure*}[!htp]
    \centering
    \begin{subfloatrow*}
        \sidesubfloat[]{
        \label{fig:dim_per_unit_linear_tinyimagenet_clean}
            \includegraphics[width=0.31\textwidth]{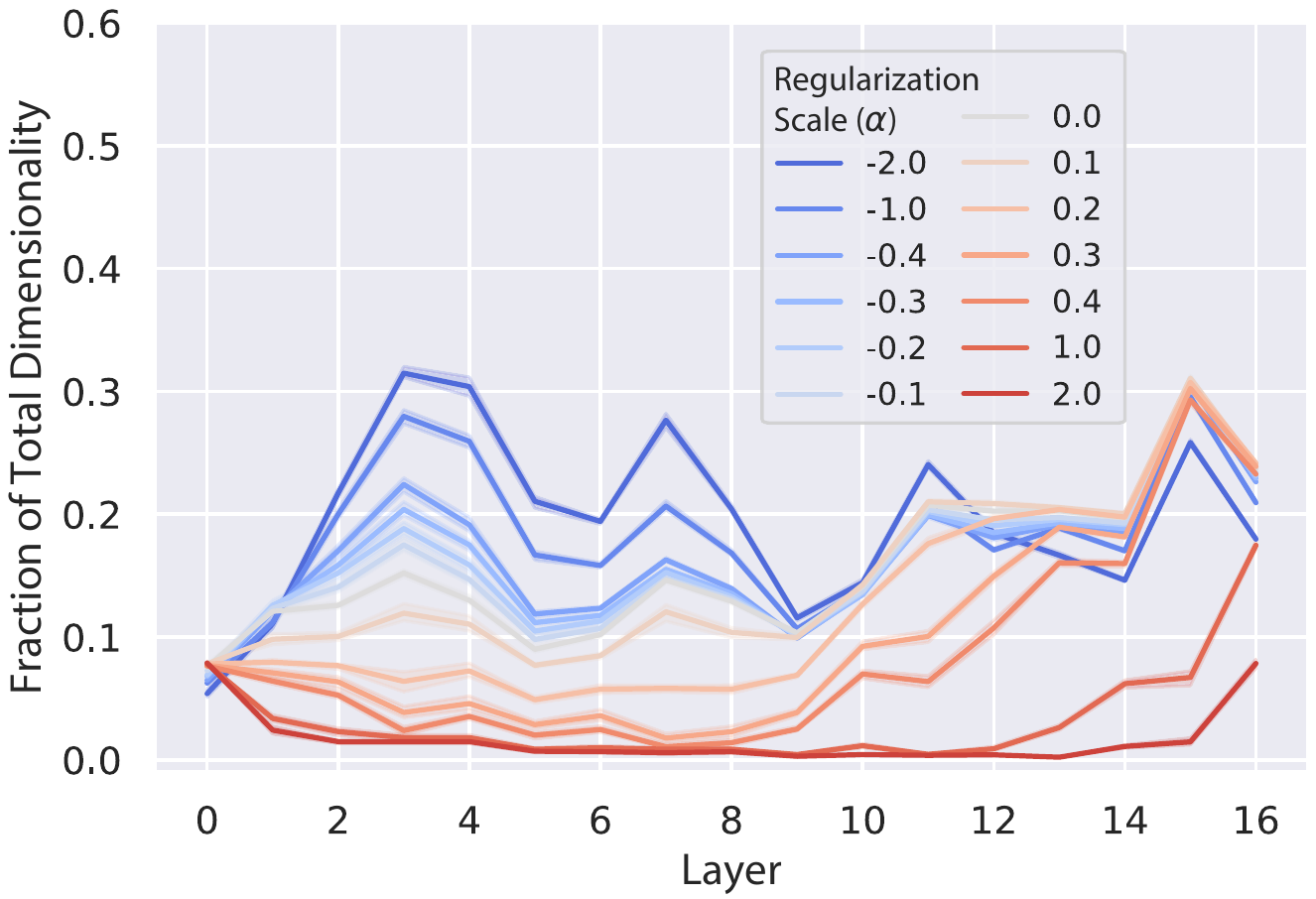}
        }
        \hspace{-6mm}
        \sidesubfloat[]{
            \label{fig:dim_per_unit_linear_tinyimagenetc_diff}
            \includegraphics[width=0.31\textwidth]{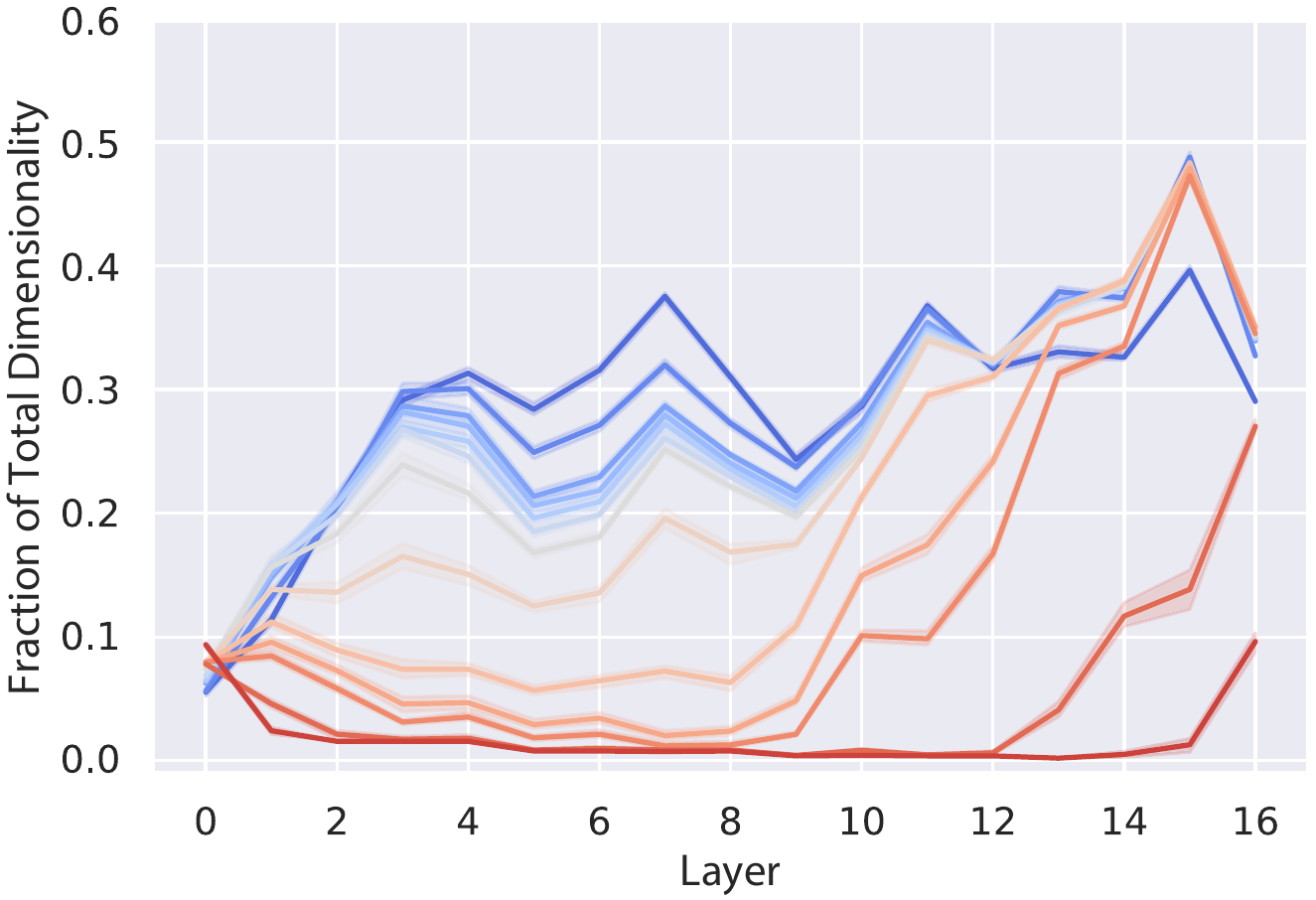}
        }        
        \hspace{1mm}
        \sidesubfloat[]{
            \label{fig:dim_per_unit_linear_tinyimagenet_adv_diff}
            \includegraphics[width=0.31\textwidth]{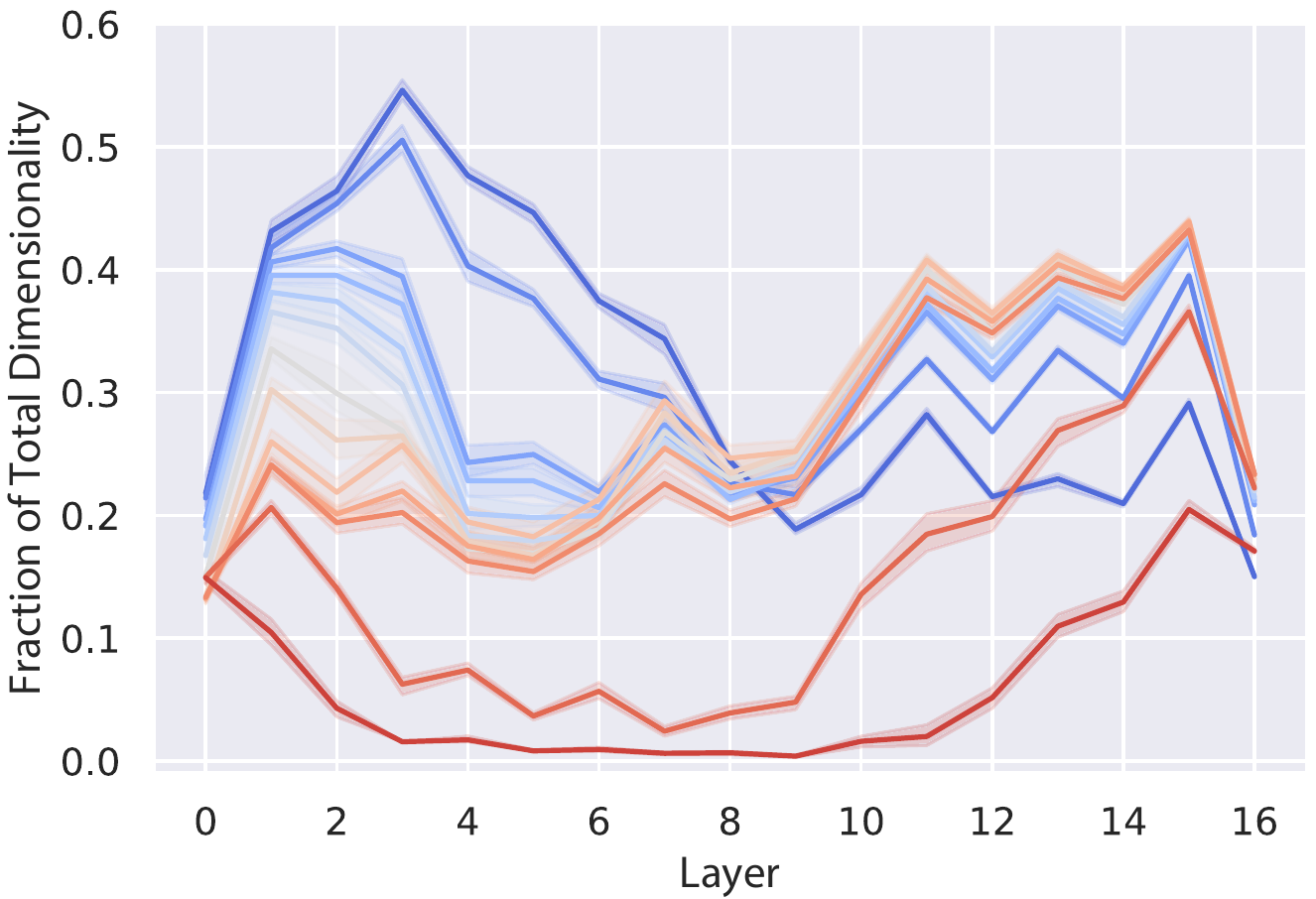}
        }
    \end{subfloatrow*}
    \vspace{-6mm}
    \caption{\textbf{Dimensionality in early layers predicts adversarial vulnerability.} (\textbf{a}) Fraction of dimensionality (y-axis; see Section \ref{sec:approach_dimensionality}) as a function of layer (x-axis). (\textbf{b}) Dimensionality of difference between clean and corrupted activations (y-axis) as a function of layer (x-axis) for Tiny ImageNetC. (\textbf{c}) Dimensionality of difference between clean and adversarial activations (y-axis) as a function of layer (x-axis). Results shown are for ResNet18 trained on Tiny ImageNet.}
    \label{fig:dim_linear}
\end{figure*}

\subsection{Dimensionality in early layers predicts adversarial vulnerability}
\label{sec:results_dimensionality}


One possible explanation for the discrepancy between class selectivity's impact on naturalistic and adversarial corruptions is that different corruption types impact representations with varying dimensionalities. For example, if only a few neurons are needed to change a network's decision, the dimensionality of the change in the representation due to corruption might be very low, as only a few units need to be modified. We thus measured dimensionality using a straightforward, linear method: we applied Principal Component Analysis (PCA) to the activation matrices of each layer in our networks and computed the number of components necessary to explain 90\% of the variance. We first examined the dimensionality of the representations of the clean test data. If the sparsity of semantic representations is reflected in dimensionality, then networks with more class selectivity should have lower-dimensional representations than networks with less class selectivity. Alternatively, if high-selectivity representations are of similar dimensionality to low-selectivity representations—though spanning different sub-regions of activation space—then dimensionality would be unaffected by class selectivity.

We found that the sparsity of a DNN's semantic representations corresponds directly to the dimensionality of those representations. Dimensionality is inversely proportional to class selectivity in early ResNet18 layers ($\leq$layer 9; Figure \ref{fig:dim_per_unit_linear_tinyimagenet_clean}), and across all of ResNet20 (Figure \ref{fig:si_dim_per_unit_linear_cifar10_clean}), indicating that the sparsity of a network's semantic representations is indeed reflected in those representations' dimensionality.

We next examined the dimensionality of perturbation-induced changes in representations by subtracting the perturbed activation matrix from the clean activation matrix and computing the dimensionality of this "difference matrix" (see Appendix \ref{sec:apx_dimensionality}). Interestingly, we found that the dimensionality of the changes in activations induced by both naturalistic (Figure \ref{fig:dim_per_unit_linear_tinyimagenetc_diff}) and adversarial perturbations (Figure \ref{fig:dim_per_unit_linear_tinyimagenet_adv_diff}) was notably higher for networks with reduced class-selectivity, again suggesting that decreasing class selectivity leads to more semantically distributed representations. 


We found that the activation changes caused by naturalistic corruptions are higher-dimensional than the representations of the clean data in both ResNet18 (compare Figures \ref{fig:dim_per_unit_linear_tinyimagenetc_diff} and \ref{fig:dim_per_unit_linear_tinyimagenet_clean}) and ResNet20 (Figures \ref{fig:si_dim_per_unit_linear_cifar10c_diff} and \ref{fig:si_dim_per_unit_linear_cifar10_clean}), and that this effect is inversely proportional to class selectivity (Figures \ref{fig:dim_per_unit_linear_tinyimagenetc_diff} and \ref{fig:si_dim_per_unit_linear_cifar10c_diff}); the increase in dimensionality from naturalistic corruptions was more pronounced in low-selectivity networks than in high-selectivity networks. These results indicate that class selectivity not only predicts the dimensionality of a representation, but also the change in dimensionality induced by a naturalistic corruption.

Notably, however, the increase in early-layer dimensionality was much larger for adversarial corruptions than naturalistic corruptions (Figure \ref{fig:dim_per_unit_linear_tinyimagenet_adv_diff}; Figure \ref{fig:si_dim_per_unit_linear_cifar10_adv_diff}). 
These results indicate that, while the changes in dimensionality induced by both naturalistic and adversarial perturbations are proportional to the dimensionality of the network's representations, these changes do not consistently project onto coding-relevant dimensions of the representations. Indeed, the larger change in early-layer dimensionality caused by adversarial perturbations likely reflects targeted projection onto coding-relevant dimensions and provides intuition as to why low-selectivity networks are more susceptible to adversarial perturbations.  



\section{Conclusion} \label{sec:conclusion}

Our results demonstrate that changes in the sparsity of semantic representations, as measured with class selectivity, provide a trade-off between robustness to naturalistic vs. adversarial perturbations: highly-distributed semantic representations confer robustness to naturalistic corruptions but result in less stable representations, causing vulnerability to adversarial perturbations. Sparse semantic representations provide the inverse trade-off: stability and adversarial robustness at the expense of vulnerability to naturalistic corruptions. The sparsity of a network's semantic representations is also reflected in the dimensionality of its activations; low class selectivity results in higher-dimensional representations than high class selectivity. We also found that the dimensionality of the difference in early-layer activations between clean and perturbed samples is larger for adversarial perturbations than for naturalistic corruptions. More generally, our results highlight a potential trade-off between robustness to adversarial vs. naturalistic perturbations, and the roles of class selectivity and representational dimensionality in mediating this effect. 
\subsubsection*{Acknowledgements} \label{sec:acknowledgements}
We would like to thank Lyndon Duong, Sho Yaida, Eric Mintun, and Rapha Gontijo Lopes for their helpful feedback, and Dan Hendrycks and Thomas Dietterich for providing the clean Tiny ImageNetC data.

\setlength{\labelsep}{\defaultlength}
\bibliography{ms}

\begin{thebibliography}{33}
\providecommand{\natexlab}[1]{#1}
\providecommand{\url}[1]{\texttt{#1}}
\expandafter\ifx\csname urlstyle\endcsname\relax
  \providecommand{\doi}[1]{doi: #1}\else
  \providecommand{\doi}{doi: \begingroup \urlstyle{rm}\Url}\fi

\bibitem[Amjad et~al.(2018)Amjad, Liu, and Geiger]{amjad_understanding_2018}
Rana~Ali Amjad, Kairen Liu, and Bernhard~C. Geiger.
\newblock Understanding {Individual} {Neuron} {Importance} {Using}
  {Information} {Theory}.
\newblock April 2018.
\newblock URL \url{https://arxiv.org/abs/1804.06679v3}.

\bibitem[Carlini and Wagner(2017{\natexlab{a}})]{carlini_adversarial_2017}
Nicholas Carlini and David Wagner.
\newblock Adversarial {Examples} {Are} {Not} {Easily} {Detected}: {Bypassing}
  {Ten} {Detection} {Methods}.
\newblock In \emph{Proceedings of the 10th {ACM} {Workshop} on {Artificial}
  {Intelligence} and {Security}}, {AISec} '17, pages 3--14, Dallas, Texas, USA,
  November 2017{\natexlab{a}}. Association for Computing Machinery.
\newblock ISBN 978-1-4503-5202-4.
\newblock \doi{10.1145/3128572.3140444}.
\newblock URL \url{https://doi.org/10.1145/3128572.3140444}.

\bibitem[Carlini and Wagner(2017{\natexlab{b}})]{carlini_towards_2017}
Nicholas Carlini and David Wagner.
\newblock Towards {Evaluating} the {Robustness} of {Neural} {Networks}.
\newblock In \emph{2017 {IEEE} {Symposium} on {Security} and {Privacy} ({SP})},
  pages 39--57, May 2017{\natexlab{b}}.
\newblock \doi{10.1109/SP.2017.49}.
\newblock ISSN: 2375-1207.

\bibitem[Dalvi et~al.(2019)Dalvi, Nortonsmith, Bau, Belinkov, Sajjad, Durrani,
  and Glass]{dalvi_neurox_2019}
Fahim Dalvi, Avery Nortonsmith, Anthony Bau, Yonatan Belinkov, Hassan Sajjad,
  Nadir Durrani, and James Glass.
\newblock {NeuroX}: {A} {Toolkit} for {Analyzing} {Individual} {Neurons} in
  {Neural} {Networks}.
\newblock \emph{Proceedings of the AAAI Conference on Artificial Intelligence},
  33\penalty0 (01):\penalty0 9851--9852, July 2019.
\newblock ISSN 2374-3468.
\newblock \doi{10.1609/aaai.v33i01.33019851}.
\newblock URL \url{https://www.aaai.org/ojs/index.php/AAAI/article/view/5063}.

\bibitem[Dhamdhere et~al.(2019)Dhamdhere, Sundararajan, and
  Yan]{dhamdhere_conductance_2019}
Kedar Dhamdhere, Mukund Sundararajan, and Qiqi Yan.
\newblock How {Important} is a {Neuron}.
\newblock In \emph{International {Conference} on {Learning} {Representations}},
  2019.
\newblock URL \url{https://openreview.net/forum?id=SylKoo0cKm}.

\bibitem[Donnelly and Roegiest(2019)]{donnelly_interpretability_2019}
Jonathan Donnelly and Adam Roegiest.
\newblock On {Interpretability} and {Feature} {Representations}: {An}
  {Analysis} of the {Sentiment} {Neuron}.
\newblock In Leif Azzopardi, Benno Stein, Norbert Fuhr, Philipp Mayr, Claudia
  Hauff, and Djoerd Hiemstra, editors, \emph{Advances in {Information}
  {Retrieval}}, Lecture {Notes} in {Computer} {Science}, pages 795--802, Cham,
  2019. Springer International Publishing.
\newblock ISBN 978-3-030-15712-8.
\newblock \doi{10.1007/978-3-030-15712-8_55}.

\bibitem[Erhan et~al.(2009)Erhan, Bengio, Courville, and
  Vincent]{erhan_visualizing_2009}
Dumitru Erhan, Yoshua Bengio, Aaron~C. Courville, and Pascal Vincent.
\newblock Visualizing {Higher}-{Layer} {Features} of a {Deep} {Network}.
\newblock 2009.

\bibitem[Fei-Fei et~al.(2015)Fei-Fei, Karpathy, and Johnson]{tiny_imagenet}
Li~Fei-Fei, Andrej Karpathy, and Justin Johnson.
\newblock Tiny imagenet visual recognition challenge, 2015.
\newblock URL \url{https://tiny-imagenet.herokuapp.com/}.

\bibitem[Gilmer et~al.(2018)Gilmer, Adams, Goodfellow, Andersen, and
  Dahl]{gilmer_motivating_2018}
Justin Gilmer, Ryan~P. Adams, Ian Goodfellow, David Andersen, and George~E.
  Dahl.
\newblock Motivating the {Rules} of the {Game} for {Adversarial} {Example}
  {Research}.
\newblock July 2018.
\newblock URL \url{https://arxiv.org/abs/1807.06732v2}.

\bibitem[Goodfellow et~al.(2015)Goodfellow, Shlens, and
  Szegedy]{goodfellow_explaining_2015}
Ian~J. Goodfellow, Jonathon Shlens, and Christian Szegedy.
\newblock Explaining and {Harnessing} {Adversarial} {Examples}.
\newblock \emph{arXiv:1412.6572 [cs, stat]}, March 2015.
\newblock URL \url{http://arxiv.org/abs/1412.6572}.
\newblock arXiv: 1412.6572.

\bibitem[He et~al.(2016)He, Zhang, Ren, and Sun]{he_resnet_2016}
Kaiming He, Xiangyu Zhang, Shaoqing Ren, and Jian Sun.
\newblock Deep residual learning for image recognition.
\newblock In \emph{Proceedings of the {IEEE} conference on computer vision and
  pattern recognition}, pages 770--778, 2016.

\bibitem[Hendrycks and Dietterich(2019)]{hendrycks_tinyimagenetc_2019}
Dan Hendrycks and Thomas~G. Dietterich.
\newblock Benchmarking {Neural} {Network} {Robustness} to {Common}
  {Corruptions} and {Surface} {Variations}.
\newblock \emph{arXiv:1807.01697 [cs, stat]}, April 2019.
\newblock URL \url{http://arxiv.org/abs/1807.01697}.
\newblock arXiv: 1807.01697.

\bibitem[Hoffman et~al.(2019)Hoffman, Roberts, and Yaida]{sho_jacobian_2019}
Judy Hoffman, Daniel~A. Roberts, and Sho Yaida.
\newblock Robust {Learning} with {Jacobian} {Regularization}.
\newblock \emph{arXiv:1908.02729 [cs, stat]}, August 2019.
\newblock URL \url{http://arxiv.org/abs/1908.02729}.
\newblock arXiv: 1908.02729.

\bibitem[Idelbayev(2020)]{github_pytorch_resnet20_2020}
Yerlan Idelbayev.
\newblock akamaster/pytorch\_resnet\_cifar10, January 2020.
\newblock URL \url{https://github.com/akamaster/pytorch_resnet_cifar10}.
\newblock original-date: 2018-01-15T09:50:56Z.

\bibitem[Karpathy et~al.(2016)Karpathy, Johnson, and
  Fei-Fei]{karpathy_visualizing_rnns_2016}
Andrej Karpathy, Justin Johnson, and Li~Fei-Fei.
\newblock Visualizing and {Understanding} {Recurrent} {Networks}.
\newblock In \emph{International {Conference} on {Learning} {Representations}},
  page~11, 2016.

\bibitem[Krizhevsky(2009)]{krizhevsky_cifar10_2009}
Alex Krizhevsky.
\newblock Learning {Multiple} {Layers} of {Features} from {Tiny} {Images}.
\newblock Technical report, 2009.

\bibitem[Kurakin et~al.(2016)Kurakin, Goodfellow, and
  Bengio]{kurakin_adversarial_2016}
Alexey Kurakin, Ian~J. Goodfellow, and Samy Bengio.
\newblock Adversarial examples in the physical world.
\newblock November 2016.
\newblock URL \url{https://openreview.net/forum?id=S1OufnIlx}.

\bibitem[Kurakin et~al.(2017)Kurakin, Goodfellow, and
  Bengio]{kurakin_adversarial_2017}
Alexey Kurakin, Ian~J. Goodfellow, and Samy Bengio.
\newblock Adversarial {Machine} {Learning} at {Scale}.
\newblock In \emph{5th {International} {Conference} on {Learning}
  {Representations}, {ICLR} 2017, {Toulon}, {France}, {April} 24-26, 2017,
  {Conference} {Track} {Proceedings}}. OpenReview.net, 2017.
\newblock URL \url{https://openreview.net/forum?id=BJm4T4Kgx}.

\bibitem[Leavitt and Morcos(2020)]{leavitt_selectivity_2020}
Matthew~L. Leavitt and Ari Morcos.
\newblock Selectivity considered harmful: evaluating the causal impact of class
  selectivity in {DNNs}.
\newblock \emph{arXiv:2003.01262 [cs, q-bio, stat]}, March 2020.
\newblock URL \url{http://arxiv.org/abs/2003.01262}.
\newblock arXiv: 2003.01262.

\bibitem[Lillian et~al.(2018)Lillian, Meyes, and Meisen]{lillian_ablation_2018}
Peter~E. Lillian, Richard Meyes, and Tobias Meisen.
\newblock Ablation of a {Robot}'s {Brain}: {Neural} {Networks} {Under} a
  {Knife}.
\newblock December 2018.
\newblock URL \url{https://arxiv.org/abs/1812.05687v2}.

\bibitem[Madry et~al.(2018)Madry, Makelov, Schmidt, Tsipras, and
  Vladu]{madry_towards_2018}
Aleksander Madry, Aleksandar Makelov, Ludwig Schmidt, Dimitris Tsipras, and
  Adrian Vladu.
\newblock Towards {Deep} {Learning} {Models} {Resistant} to {Adversarial}
  {Attacks}.
\newblock February 2018.
\newblock URL \url{https://openreview.net/forum?id=rJzIBfZAb}.

\bibitem[Morcos et~al.(2018)Morcos, Barrett, Rabinowitz, and
  Botvinick]{morcos_single_directions_2018}
Ari~S. Morcos, David G.~T. Barrett, Neil~C. Rabinowitz, and Matthew Botvinick.
\newblock On the importance of single directions for generalization.
\newblock In \emph{International {Conference} on {Learning} {Representations}},
  2018.
\newblock URL \url{https://openreview.net/forum?id=r1iuQjxCZ}.

\bibitem[Novak et~al.(2018)Novak, Bahri, Abolafia, Pennington, and
  Sohl-Dickstein]{novak_sensitivity_2018}
Roman Novak, Yasaman Bahri, Daniel~A. Abolafia, Jeffrey Pennington, and Jascha
  Sohl-Dickstein.
\newblock Sensitivity and {Generalization} in {Neural} {Networks}: an
  {Empirical} {Study}.
\newblock \emph{arXiv:1802.08760 [cs, stat]}, June 2018.
\newblock URL \url{http://arxiv.org/abs/1802.08760}.
\newblock arXiv: 1802.08760.

\bibitem[Olah et~al.(2020)Olah, Cammarata, Schubert, Goh, Petrov, and
  Carter]{olah_zoom_2020}
Chris Olah, Nick Cammarata, Ludwig Schubert, Gabriel Goh, Michael Petrov, and
  Shan Carter.
\newblock Zoom {In}: {An} {Introduction} to {Circuits}.
\newblock \emph{Distill}, 5\penalty0 (3):\penalty0 e00024.001, March 2020.
\newblock ISSN 2476-0757.
\newblock \doi{10.23915/distill.00024.001}.
\newblock URL \url{https://distill.pub/2020/circuits/zoom-in}.

\bibitem[Paszke et~al.(2019)Paszke, Gross, Massa, Lerer, Bradbury, Chanan,
  Killeen, Lin, Gimelshein, Antiga, Desmaison, Kopf, Yang, DeVito, Raison,
  Tejani, Chilamkurthy, Steiner, Fang, Bai, and Chintala]{paszke_pytorch_2019}
Adam Paszke, Sam Gross, Francisco Massa, Adam Lerer, James Bradbury, Gregory
  Chanan, Trevor Killeen, Zeming Lin, Natalia Gimelshein, Luca Antiga, Alban
  Desmaison, Andreas Kopf, Edward Yang, Zachary DeVito, Martin Raison, Alykhan
  Tejani, Sasank Chilamkurthy, Benoit Steiner, Lu~Fang, Junjie Bai, and Soumith
  Chintala.
\newblock {PyTorch}: {An} {Imperative} {Style}, {High}-{Performance} {Deep}
  {Learning} {Library}.
\newblock In H.~Wallach, H.~Larochelle, A.~Beygelzimer,
  F.~d{\textbackslash}textquotesingle Alché-Buc, E.~Fox, and R.~Garnett,
  editors, \emph{Advances in {Neural} {Information} {Processing} {Systems} 32},
  pages 8024--8035. Curran Associates, Inc., 2019.
\newblock URL
  \url{http://papers.neurips.cc/paper/9015-pytorch-an-imperative-style-high-performance-deep-learning-library.pdf}.

\bibitem[Rifai et~al.(2011)Rifai, Vincent, Muller, Glorot, and
  Bengio]{rifai_contractive_2011}
Salah Rifai, Pascal Vincent, Xavier Muller, Xavier Glorot, and Yoshua Bengio.
\newblock Contractive auto-encoders: explicit invariance during feature
  extraction.
\newblock In \emph{Proceedings of the 28th {International} {Conference} on
  {International} {Conference} on {Machine} {Learning}}, {ICML}'11, pages
  833--840, Bellevue, Washington, USA, June 2011. Omnipress.
\newblock ISBN 978-1-4503-0619-5.

\bibitem[Sokolic et~al.(2017)Sokolic, Giryes, Sapiro, and
  Rodrigues]{sokolic_robust_2017}
Jure Sokolic, Raja Giryes, Guillermo Sapiro, and Miguel R.~D. Rodrigues.
\newblock Robust {Large} {Margin} {Deep} {Neural} {Networks}.
\newblock \emph{IEEE Transactions on Signal Processing}, 65\penalty0
  (16):\penalty0 4265--4280, August 2017.
\newblock ISSN 1053-587X, 1941-0476.
\newblock \doi{10.1109/TSP.2017.2708039}.
\newblock URL \url{http://arxiv.org/abs/1605.08254}.
\newblock arXiv: 1605.08254.

\bibitem[Szegedy et~al.(2013)Szegedy, Zaremba, Sutskever, Bruna, Erhan,
  Goodfellow, and Fergus]{szegedy_intriguing_2013}
Christian Szegedy, Wojciech Zaremba, Ilya Sutskever, Joan Bruna, Dumitru Erhan,
  Ian Goodfellow, and Rob Fergus.
\newblock Intriguing properties of neural networks.
\newblock December 2013.
\newblock URL \url{https://arxiv.org/abs/1312.6199v4}.

\bibitem[Vasiljevic et~al.(2016)Vasiljevic, Chakrabarti, and
  Shakhnarovich]{vasiljevic_examining_2016}
Igor Vasiljevic, Ayan Chakrabarti, and Gregory Shakhnarovich.
\newblock Examining the {Impact} of {Blur} on {Recognition} by {Convolutional}
  {Networks}.
\newblock November 2016.
\newblock URL \url{https://arxiv.org/abs/1611.05760v2}.

\bibitem[Virtanen et~al.(2019)Virtanen, Gommers, Oliphant, Haberland, Reddy,
  Cournapeau, Burovski, Peterson, Weckesser, Bright, van~der Walt, Brett,
  Wilson, Millman, Mayorov, Nelson, Jones, Kern, Larson, Carey, Polat, Feng,
  Moore, VanderPlas, Laxalde, Perktold, Cimrman, Henriksen, Quintero, Harris,
  Archibald, Ribeiro, Pedregosa, van Mulbregt, and
  Contributors]{virtanen_scipy_2019}
Pauli Virtanen, Ralf Gommers, Travis~E. Oliphant, Matt Haberland, Tyler Reddy,
  David Cournapeau, Evgeni Burovski, Pearu Peterson, Warren Weckesser, Jonathan
  Bright, Stéfan~J. van~der Walt, Matthew Brett, Joshua Wilson, K.~Jarrod
  Millman, Nikolay Mayorov, Andrew R.~J. Nelson, Eric Jones, Robert Kern, Eric
  Larson, C.~J. Carey, Ilhan Polat, Yu~Feng, Eric~W. Moore, Jake VanderPlas,
  Denis Laxalde, Josef Perktold, Robert Cimrman, Ian Henriksen, E.~A. Quintero,
  Charles~R. Harris, Anne~M. Archibald, Antônio~H. Ribeiro, Fabian Pedregosa,
  Paul van Mulbregt, and SciPy 1~0 Contributors.
\newblock {SciPy} 1.0--{Fundamental} {Algorithms} for {Scientific} {Computing}
  in {Python}.
\newblock \emph{arXiv:1907.10121 [physics]}, July 2019.
\newblock URL \url{http://arxiv.org/abs/1907.10121}.
\newblock arXiv: 1907.10121.

\bibitem[Waskom et~al.(2017)Waskom, Botvinnik, O'Kane, Hobson, Lukauskas,
  Gemperline, Augspurger, Halchenko, Cole, Warmenhoven, Ruiter, Pye, Hoyer,
  Vanderplas, Villalba, Kunter, Quintero, Bachant, Martin, Meyer, Miles, Ram,
  Yarkoni, Williams, Evans, Fitzgerald, {Brian}, Fonnesbeck, Lee, and
  Qalieh]{waskom_seaborn_2017}
Michael Waskom, Olga Botvinnik, Drew O'Kane, Paul Hobson, Saulius Lukauskas,
  David~C. Gemperline, Tom Augspurger, Yaroslav Halchenko, John~B. Cole, Jordi
  Warmenhoven, Julian~de Ruiter, Cameron Pye, Stephan Hoyer, Jake Vanderplas,
  Santi Villalba, Gero Kunter, Eric Quintero, Pete Bachant, Marcel Martin, Kyle
  Meyer, Alistair Miles, Yoav Ram, Tal Yarkoni, Mike~Lee Williams, Constantine
  Evans, Clark Fitzgerald, {Brian}, Chris Fonnesbeck, Antony Lee, and Adel
  Qalieh.
\newblock mwaskom/seaborn: v0.8.1 ({September} 2017), September 2017.
\newblock URL \url{https://doi.org/10.5281/zenodo.883859}.

\bibitem[Zeiler and Fergus(2014)]{zeiler_deconvolution_2014}
Matthew~D. Zeiler and Rob Fergus.
\newblock Visualizing and {Understanding} {Convolutional} {Networks}.
\newblock In David Fleet, Tomas Pajdla, Bernt Schiele, and Tinne Tuytelaars,
  editors, \emph{Computer {Vision} – {ECCV} 2014}, pages 818--833, Cham,
  2014. Springer International Publishing.
\newblock ISBN 978-3-319-10590-1.

\bibitem[Zheng et~al.(2016)Zheng, Song, Leung, and
  Goodfellow]{zheng_improving_2016}
Stephan Zheng, Yang Song, Thomas Leung, and Ian Goodfellow.
\newblock Improving the {Robustness} of {Deep} {Neural} {Networks} via
  {Stability} {Training}.
\newblock April 2016.
\newblock URL \url{https://arxiv.org/abs/1604.04326v1}.

\end{thebibliography}
\bibliographystyle{plainnat}
\setlength{\labelsep}{\adjlength}

\clearpage
\onecolumn
\renewcommand\thesection{\Alph{section}}
\setcounter{section}{0}
\renewcommand\thefigure{A\arabic{figure}}
\setcounter{figure}{0}
\phantomsection

\section{Appendix}
\label{sec:appendix}

\subsection{Detailed approach} \label{appendix:models_training_etc}

Unless otherwise noted: all experimental results were derived from the corrupted or adversarial test set with the parameters from the epoch that achieved the highest clean validation set accuracy over the training epochs; 20 replicates with different random seeds were run for each hyperparameter set; error bars and shaded regions denote bootstrapped 95\% confidence intervals; selectivity regularization was not applied to the final (output) layer, nor was the final layer included in any of our analyses.

\subsubsection{Models}
\label{sec:apx_models}
All models were trained using stochastic gradient descent (SGD) with momentum = 0.9 and weight decay = 0.0001. The maxpool layer after the first batchnorm layer in ResNet18 (see \citet{he_resnet_2016}) was removed because of the smaller size of Tiny Imagenet images compared to standard ImageNet images (64x64 vs. 256x256, respectively). ResNet18 were trained for 90 epochs with a minibatch size of 4096 samples with a learning rate of 0.1, multiplied (annealed) by 0.1 at epochs 35, 50, 65, and 80. .

ResNet20 (code modified from \citet{github_pytorch_resnet20_2020}) were trained for 200 epochs using a minibatch size of 256 samples and a learning rate of 0.1, annealed by 0.1 at epochs 100 and 150.

\subsubsection{Datasets}
\label{sec:apx_data}
Tiny Imagenet \citep{tiny_imagenet} consists of 500 training images and 50 images for each of its 200 classes. We used the validation set for testing and created a new validation set by taking 50 images per class from the training set, selected randomly for each seed. We split the 50k CIFAR10 training samples into a 45k sample training set and a 5k validation set, similar to our approach with Tiny Imagenet.

All experimental results were derived from the test set with the parameters from the epoch that achieved the highest validation set accuracy over the training epochs. 20 replicates with different random seeds were run for each hyperparameter set. Selectivity regularization was not applied to the final (output) layer, nor was the final layer included any of our analyses.

CIFAR10C consists of a dataset in which 19 different naturalistic corruptions have been applied to the CIFAR10 test set at 5 different levels of severity. Tiny ImageNetC also has 5 levels of corruption severity, but consists of 15 corruptions.

We would like to note that Tiny ImageNetC does not use the Tiny ImageNet test data. While the two datasets were created using the same data generation procedure—cropping and scaling images from the same 200 ImageNet classes—they differ in the specific ImageNet images they use. It is possible that the images used to create Tiny ImageNetC are out-of-distribution with regards to the Tiny ImageNet training data, in which case our results from testing on Tiny ImageNetC actually underestimate the corruption robustness of our networks. The creators of Tiny ImageNetC kindly provided the clean (uncorrupted) Tiny ImageNetC data necessary for the dimensionality analysis, which relies on matches corrupted and clean data samples.

\subsubsection{Quantifying and regularizing class selectivity}
\label{sec:apx_class_selectivity}

We use the same approach and terminology as \citep{leavitt_selectivity_2020}. A unit's class selectivity index is calculated as follows: For a single convolutional feature map (which we refer to as a "unit"), the activation in response to a single sample was averaged across all elements of the filter map. The class-conditional mean activation (i.e. the mean activation for each class) was then calculated across all samples in the clean test set, and we then compute:

\begin{equation}
    selectivity = \frac{\mu_{max} - \mu_{-max}}{\mu_{max} + \mu_{-max}}
\end{equation}

where $\mu_{max}$ is the largest class-conditional mean activation and $\mu_{-max}$ is the mean response to the remaining (i.e. non-$\mu_{max}$) classes. The selectivity index ranges from 0 to 1. A unit with identical average activity for all classes would have a selectivity of 0, and a unit that only responds to a single class would have a selectivity of 1.

We regularize for or against class selectivity by using the following loss function:

\begin{equation}
    loss = -\sum_{c}^{C} {y_{c} \cdotp \log(\hat{y_{c}})} - \alpha\mu_{SI}
\end{equation}

The left-hand term in the loss function is the traditional cross-entropy between the softmax of the output units and the true class labels; $c$ is the class index; $C$ is the number of classes; $y_c$ is the true class label; $\hat{y_c}$ is the predicted class probability. We refer to the right-hand component of the loss function ($-\alpha\mu_{SI}$) as the class selectivity regularizer. The regularizer comprises two terms: the network selectivity, $\mu_{SI}$:

\begin{equation}
    \mu_{SI} = \frac{1}{L} \sum_{l}^{L} \frac{1}{U} {\sum_{u}^{U} {SI_{l,u}}}
\end{equation}

\ari{be careful with the vspaces, the top of the $\Sigma$ was overlapping text in eq 3}

where $l$ is a convolutional layer, $L$ is number of layers, $u$ is a unit (i.e. feature map), $U$ is the number of units in a given layer, and $SI_u$ is the class selectivity index of unit $u$. We obtain the network selectivity by computing the selectivity index for each unit in a layer, then computing the mean selectivity index for units within a layer, then computing the mean selectivity index across layers. Computing the within-layer mean before computing the across-layer mean mitigates the biases induced by the larger numbers of units in deeper layers. The regularizer's remaining term is $\alpha$, the regularizer scale. Negative values of $\alpha$ discourage class selectivity in individual units, while positive values promote it. The magnitude of $\alpha$ controls the contribution of the network selectivity to the overall loss. During training, the class selectivity index was computed for each minibatch. For the analyses presented here, the class selectivity index was computed across the entire clean test set.

\subsubsection{Adversarial attacks}
\label{sec:apx_adv}
For the PGD attack, we used a step size of 0.0001.

\subsubsection{Quantifying dimensionality}
\label{sec:apx_dimensionality}

As mentioned in Approach \ref{sec:approach_dimensionality}, we quantified the dimensionality of a layer's representations by applying PCA to the layer's activation matrix for the clean test data (or the difference between the clean test activations and the perturbed activations) and counting the number of dimensions necessary to explain 90\% of the variance. For each perturbation type (adversarial attack and naturalistic corruption), we chose a representative perturbation. For the adversarial attack, we chose PGD with 25 steps. For the naturalistic corruptions, we chose brightness at severity 3 for Tiny ImageNetC, and motion blur at severity 3 for CIFAR10C. 



\subsubsection{Software}
Experiments were conducted using PyTorch \citep{paszke_pytorch_2019}, analyzed using the SciPy ecosystem \citep{virtanen_scipy_2019}, and visualized using Seaborn \citep{waskom_seaborn_2017}. 

\clearpage
\subsection{Supplementary Results}
\begin{figure*}[!htbp]
    \centering
    \begin{subfloatrow*}
        \hspace{6mm}
        \sidesubfloat[]{
        \label{fig:si_sel_reg_resnet18}
            \includegraphics[width=0.42\textwidth]{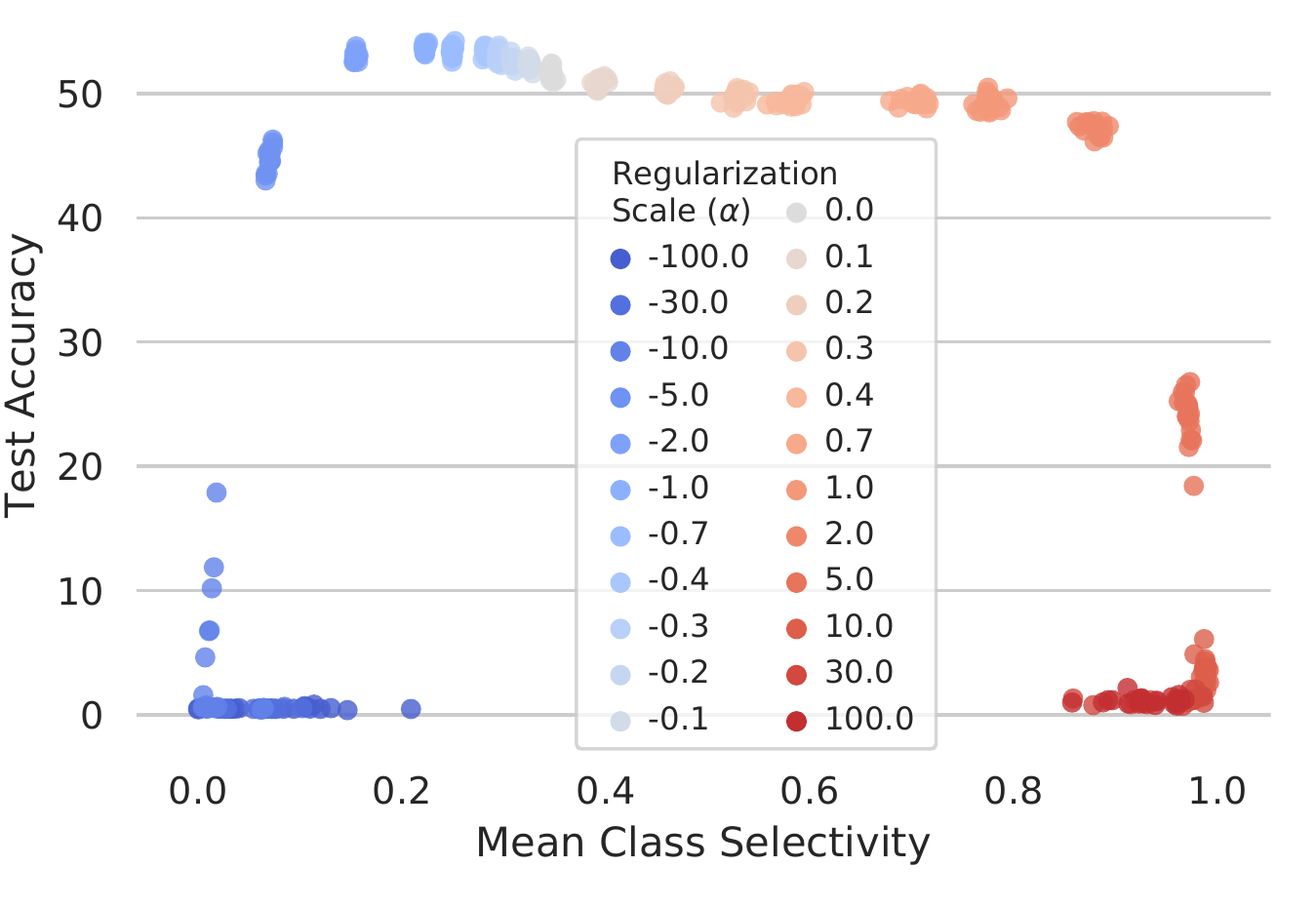}
        }
        \hspace{6mm}
        \sidesubfloat[]{
            \label{fig:si_sel_reg_resnet20}
            \includegraphics[width=0.42\textwidth]{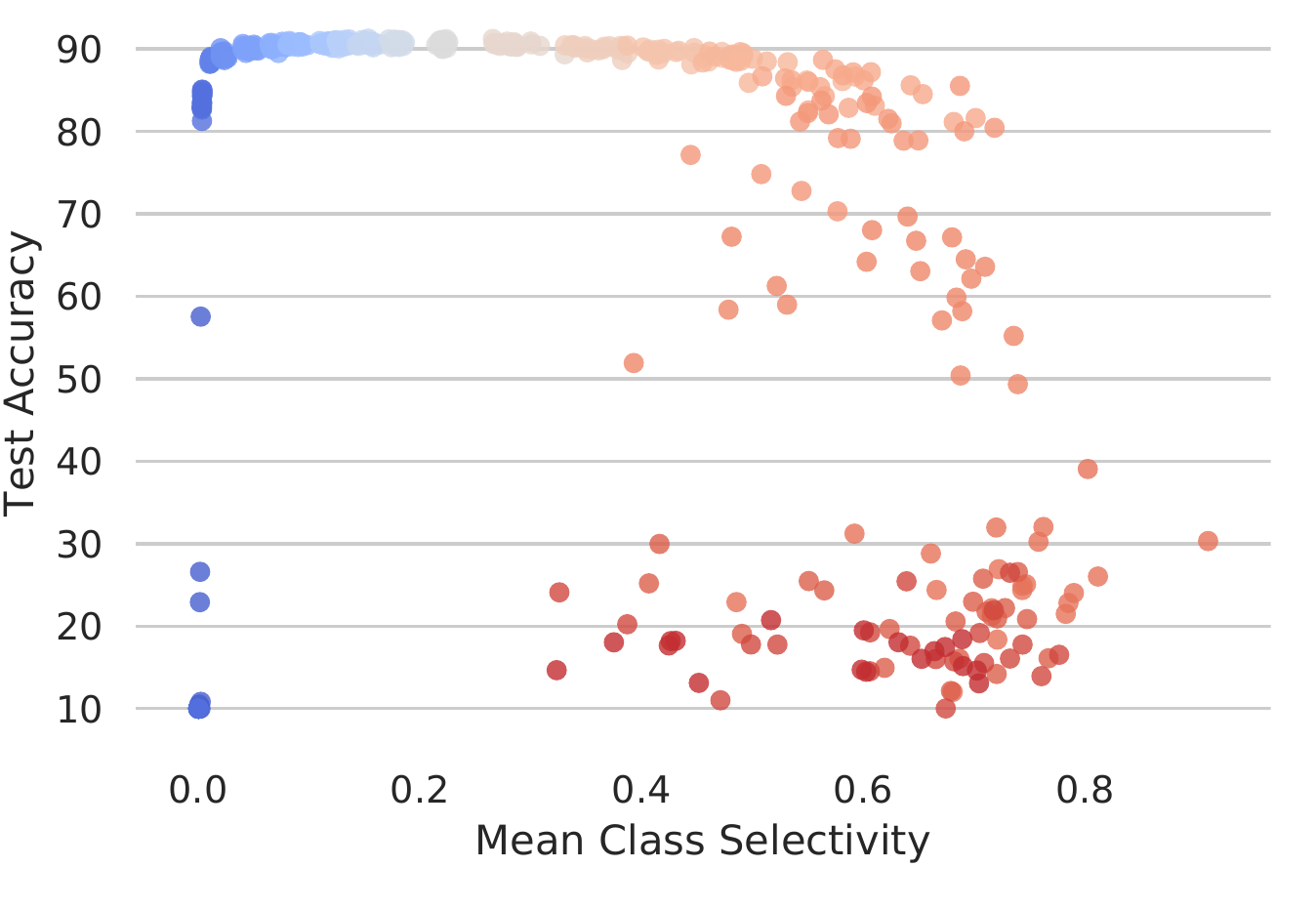}
        }
    \end{subfloatrow*}
    \vspace{-8mm}
    \caption{\textbf{Effects of class selectivity regularization on test accuracy.} (\textbf{a}) Test accuracy (y-axis) as a function of mean class selectivity (x-axis) for ResNet18 trained on Tiny ImageNet. $\alpha$ denotes the sign and intensity of class selectivity regularization. Negative $\alpha$ lowers selectivity, positive $\alpha$ increases selectivity,and the magnitude of $\alpha$ changes the strength of the effect. (\textbf{b}) Same as (\textbf{a}), but for ResNet20 trained on CIFAR10.}
    \label{fig:si_sel_reg}
\end{figure*}

\begin{figure*}[!htp]
    \centering
    \includegraphics[width=1\textwidth]{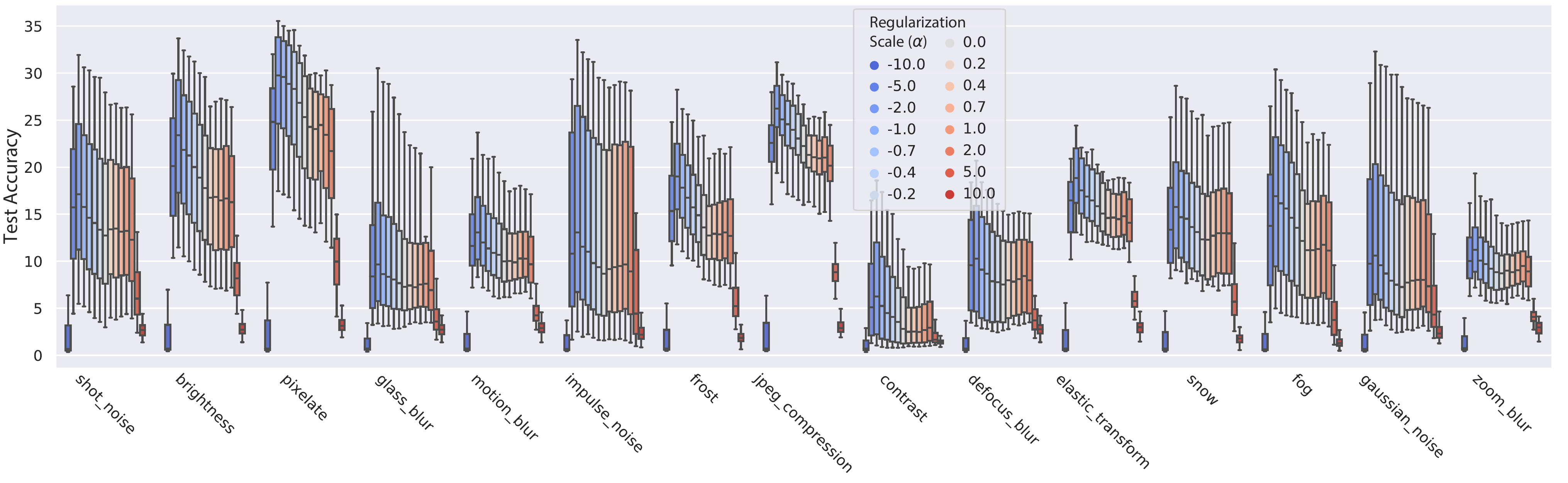}
    \vspace{-8mm}
    \caption{\textbf{Reduced class selectivity consistently improves corruption-robustness in ResNet18.} Mean test accuracy (y-axis) for each Tiny ImageNetC corruption (x-axis) and regularization scale ($\alpha$, color). Mean is computed across all perturbation intensities.}
    \label{fig:si_accuracy_every_corruption_tinyimagenetc}
\end{figure*}

\begin{figure*}[!htp]
    \centering
    \includegraphics[width=0.33\textwidth]{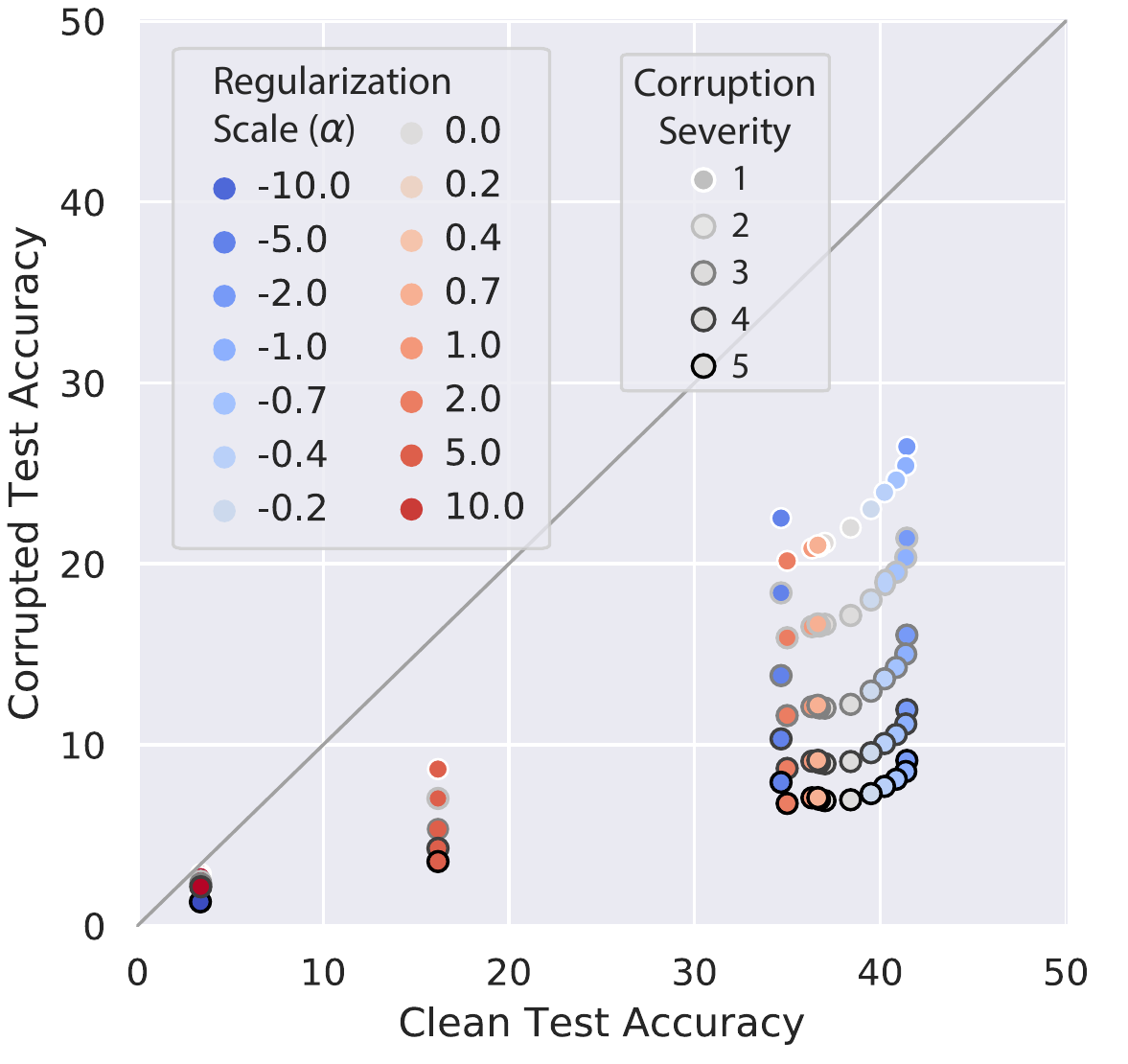}
    \vspace{-5mm}
    \caption{\textbf{Trade-off between clean and corrupted test accuracy in ResNet18 tested on Tiny ImageNetC.} Clean test accuracy (x-axis) vs. corrupted test accuracy (y-axis) for different corruption severities (border color) and regularization scales ($\alpha$, fill color). Mean is computed across all corruption types.}
    \label{fig:si_clean_corrupted_tradeoff_tinyimagenetc}
\end{figure*}

\clearpage

\subsection{Results for ResNet20 trained on CIFAR10C}
\begin{figure*}[!htp]
    \centering
    \begin{subfloatrow*}
        \sidesubfloat[]{
        \label{fig:si_acc_abs_mean_cifar10c}
            \includegraphics[width=0.30\textwidth]{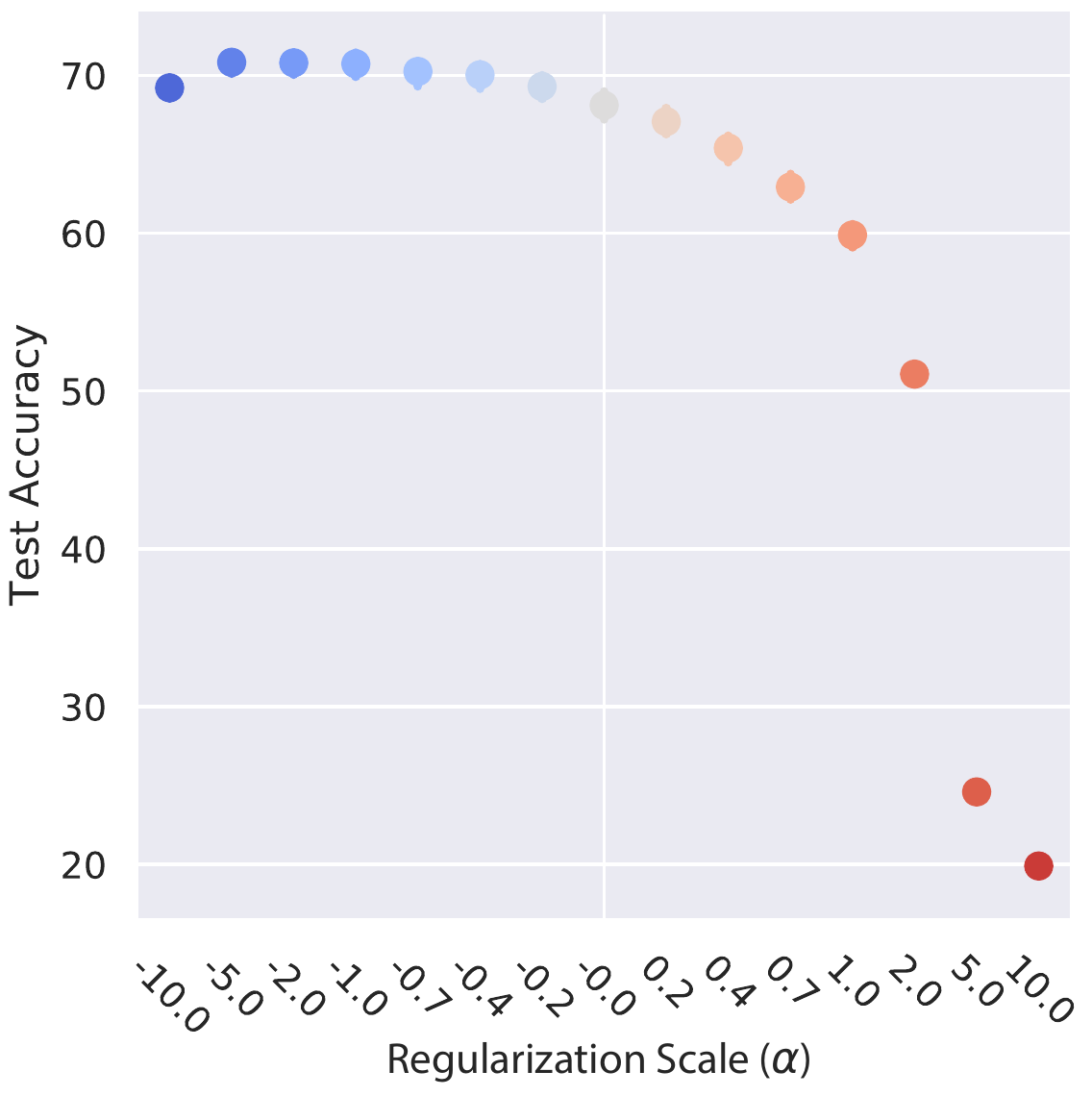}
        }
        \hspace{-7mm}
        \sidesubfloat[]{
            \label{fig:si_acc_rel_mean_cifar10c}
            \includegraphics[width=0.33\textwidth]{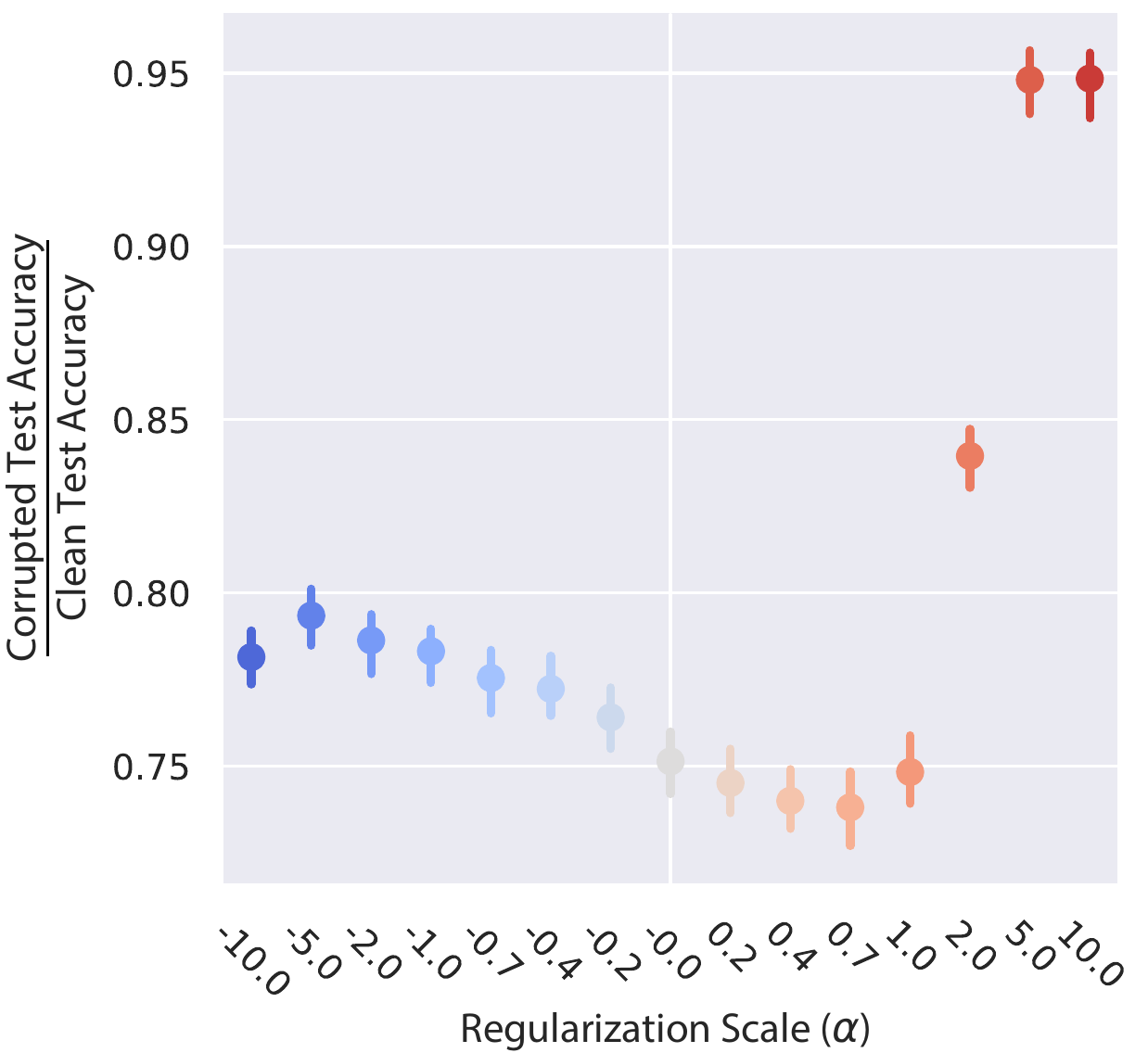}
        }
        \sidesubfloat[]{
            \label{fig:si_perturbation_types_cifar10c}
            \includegraphics[width=0.32\textwidth]{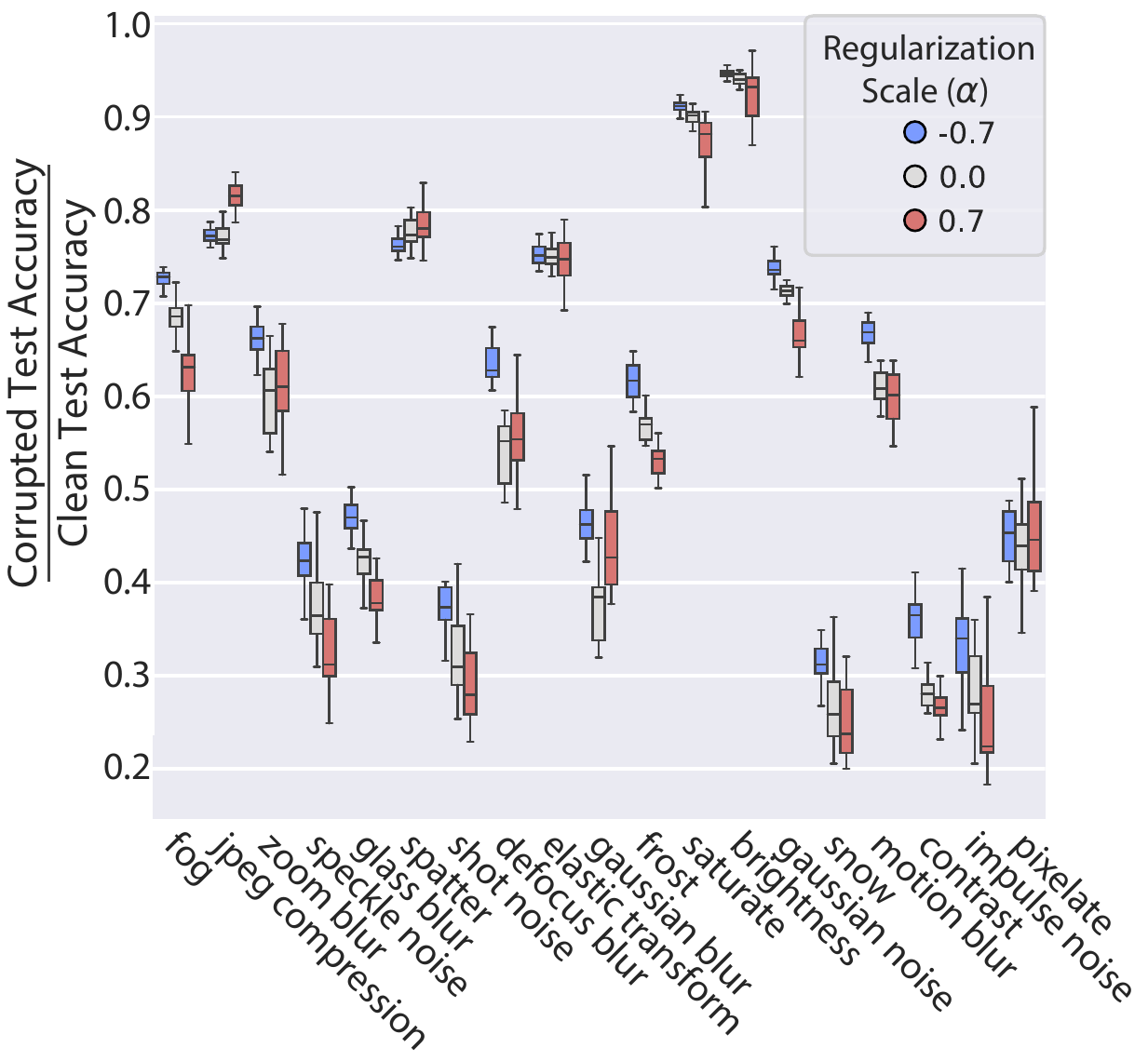}
        }
    \end{subfloatrow*}
    \vspace{-5mm}
    \caption{\textbf{Reducing class selectivity confers corruption robustness in ResNet20 trained on CIFAR10.} (\textbf{a}) Mean test accuracy across all corruptions and severities (y-axis) as a function of class selectivity regularization scale ($\alpha$; x-axis). Negative $\alpha$ lowers selectivity, positive $\alpha$ increases selectivity, and the magnitude of $\alpha$ changes the strength of the effect. (\textbf{b}) Corrupted test accuracy normalized by clean test accuracy (y-axis) as a function of $\alpha$ (x-axis). (\textbf{c}) Normalized test accuracy (y-axis) for all 19 CIFAR10C corruption types (x-axis) for three example values of $\alpha$.}
    \label{fig:si_acc_cifar10c}
\end{figure*}

\begin{figure*}[!htbp]
    \centering
    \includegraphics[width=1\textwidth]{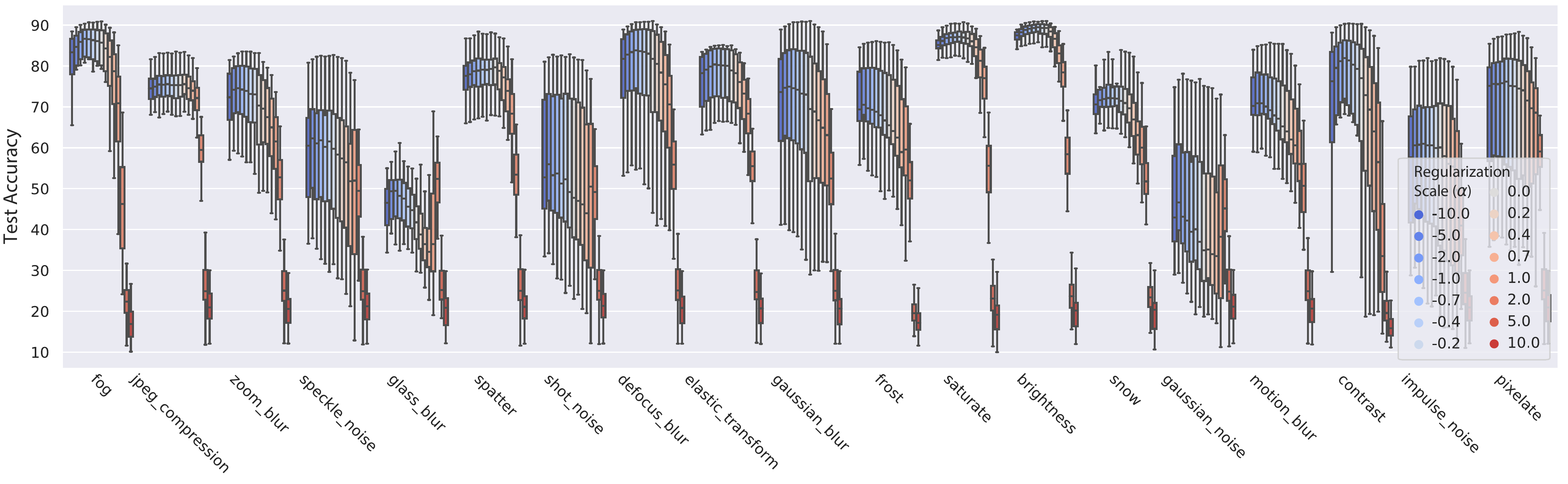}
    \vspace{-8mm}
    \caption{\textbf{Reduced class selectivity consistently improves corruption-robustness in ResNet20 trained on CIFAR10.} Mean test accuracy (y-axis) for each CIFAR10C corruption (x-axis) and regularization scale ($\alpha$, color). Mean is computed across all perturbation severities.}
    \label{fig:si_accuracy_every_corruption_cifar10c}
\end{figure*}

\begin{figure*}[!htbp]
    \centering
    \includegraphics[width=0.33\textwidth]{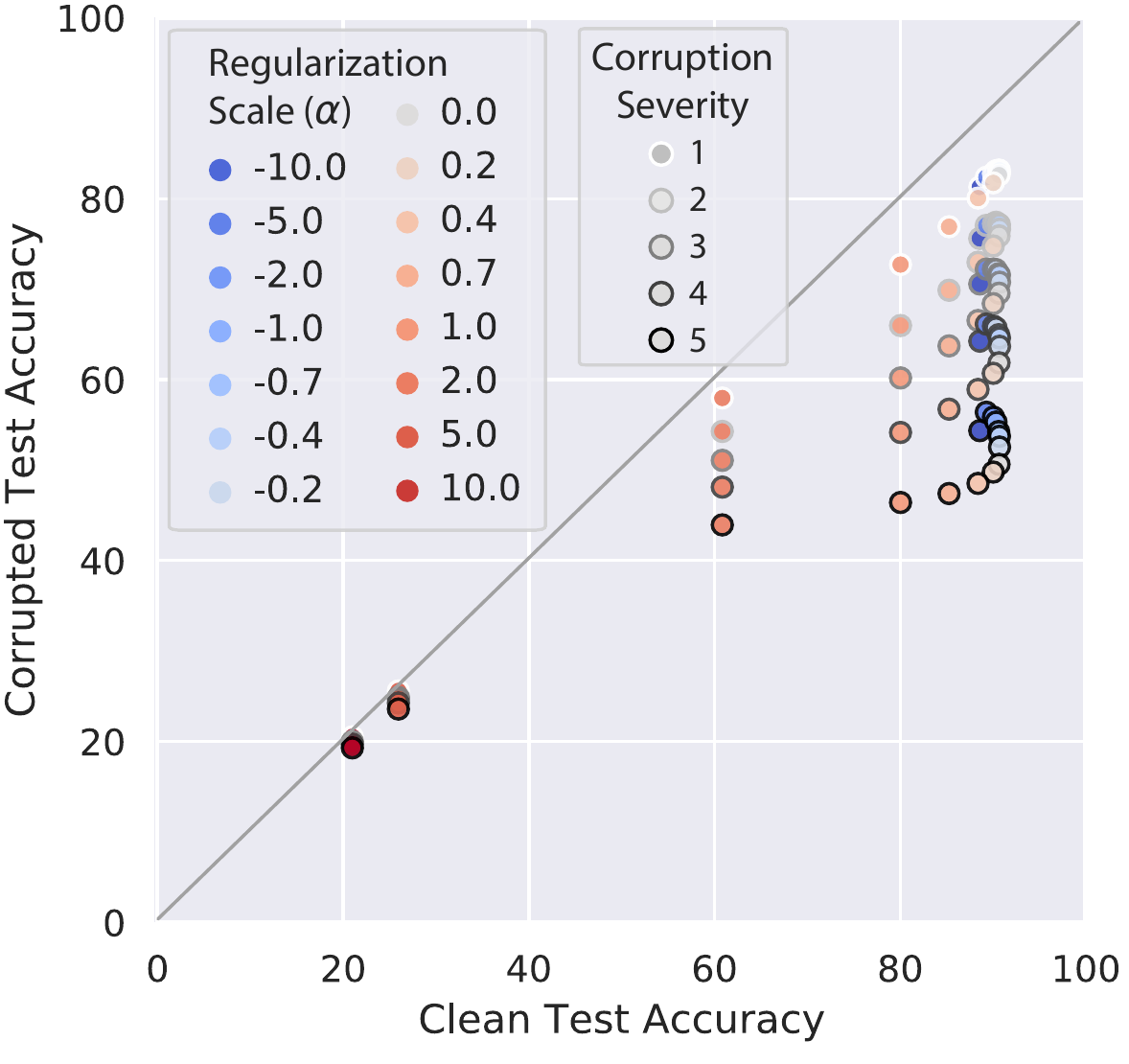}
    \vspace{-5mm}
    \caption{\textbf{Trade-off between clean and corrupted test accuracy in ResNet20 tested on CIFAR10C.} Clean test accuracy (x-axis) vs. corrupted test accuracy (y-axis) for different corruption severities (border color) and regularization scales ($\alpha$, fill color). Mean is computed across all corruption types.}
    \label{fig:si_clean_corrupted_tradeoff_cifar10c}
\end{figure*}

\begin{figure*}[!htbp]
    \centering
    \begin{subfloatrow*}
        \sidesubfloat[]{
        \label{fig:si_fgsm_resnet20}
            \includegraphics[width=0.3\textwidth]{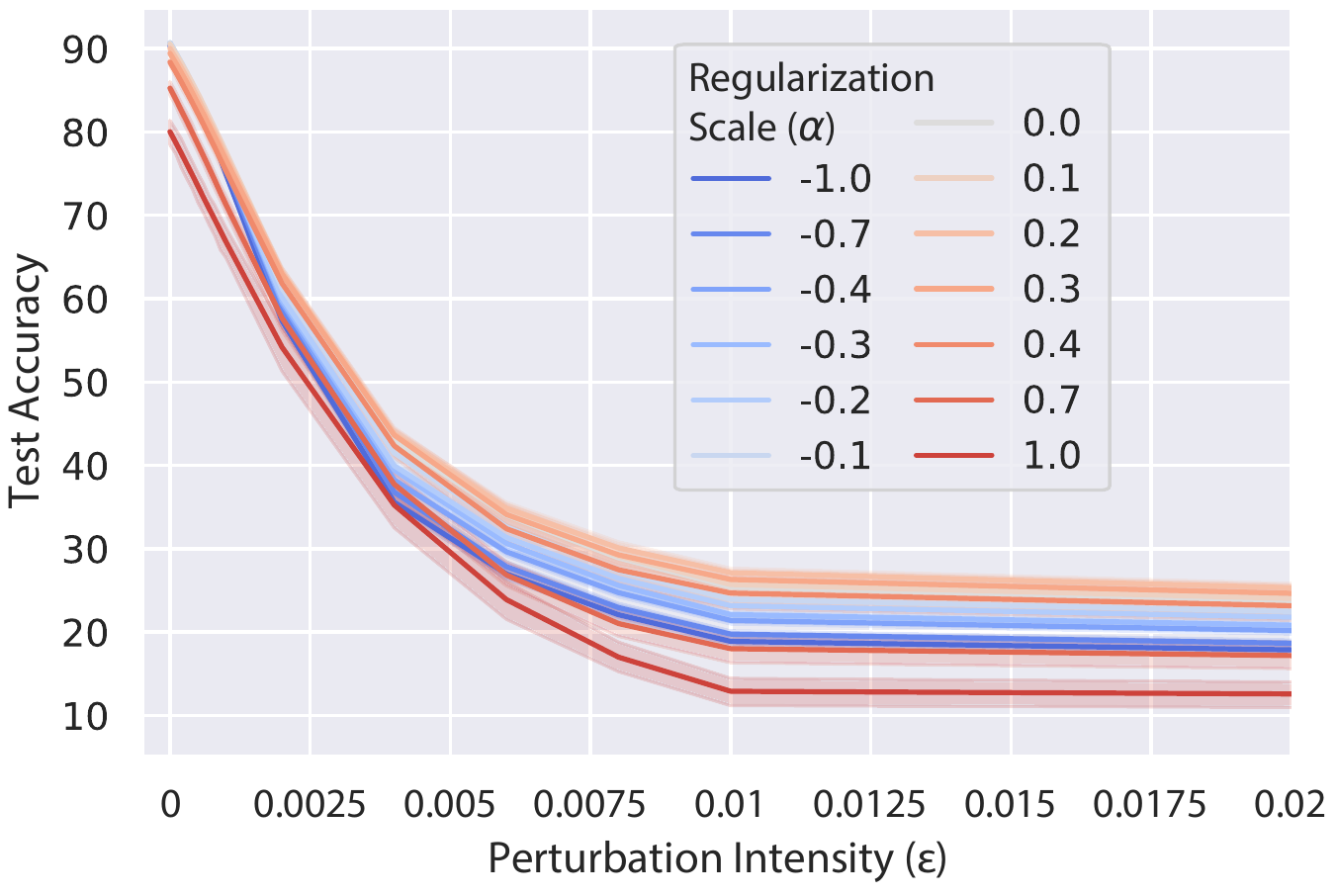}
        }
        \hspace{-4mm}
        \sidesubfloat[]{
            \label{fig:si_pgd_resnet20}
            \includegraphics[width=0.3\textwidth]{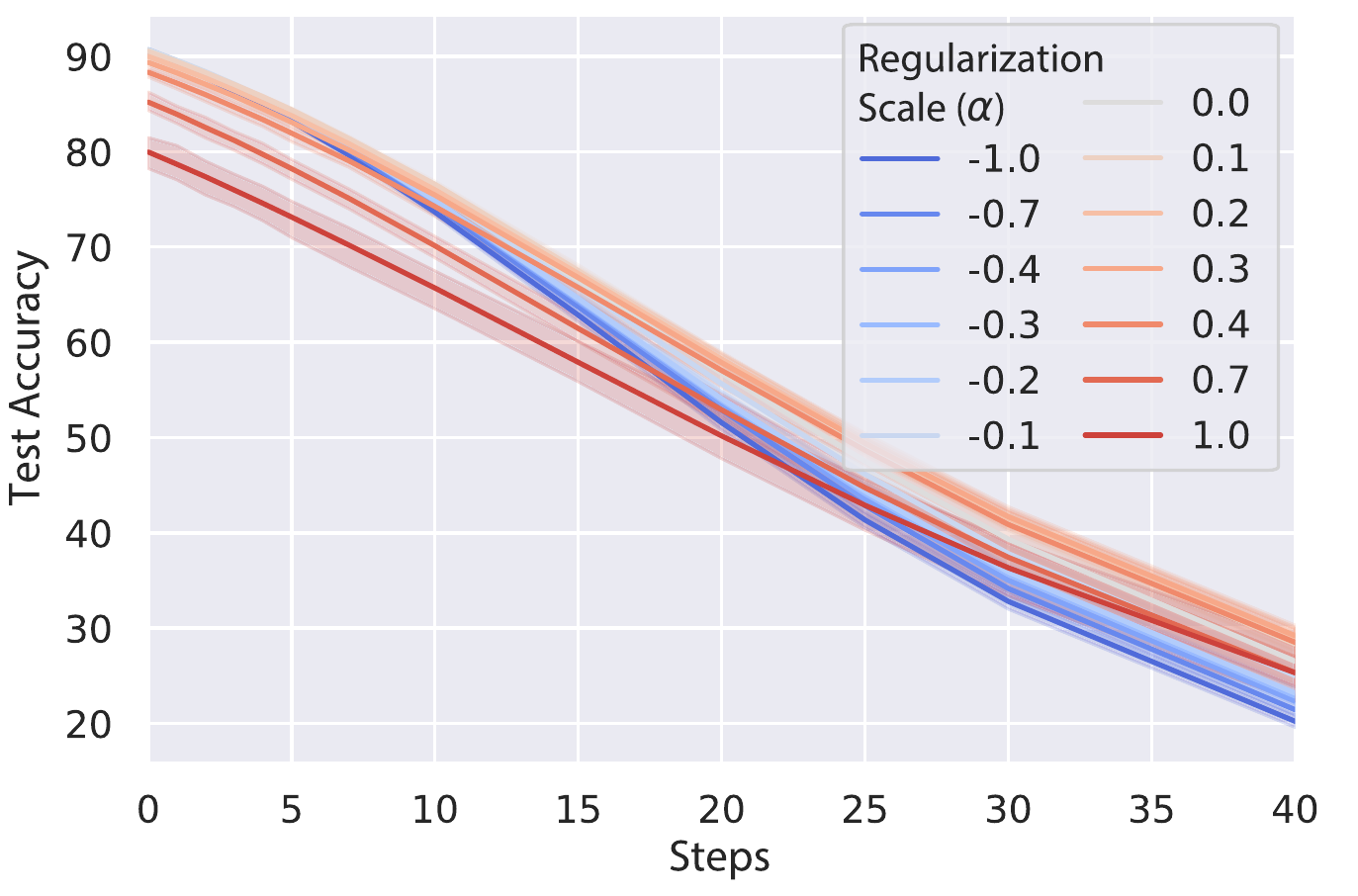}
        }
        \hspace{3mm}
        \sidesubfloat[]{
            \label{fig:si_jacobian_resnet20}
            \includegraphics[width=0.29\textwidth]{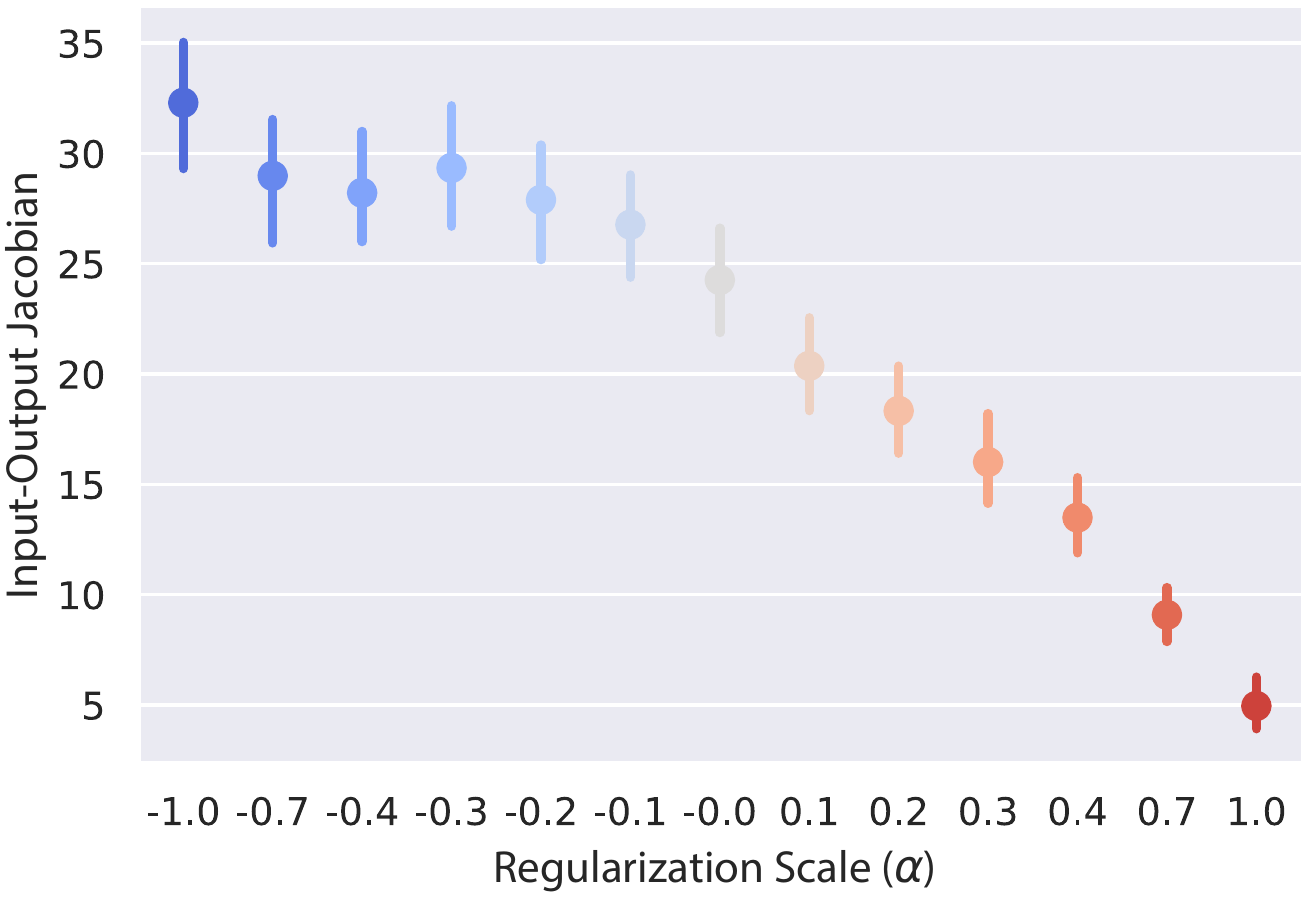}
        }
    \end{subfloatrow*}
    \vspace{-5mm}
    \caption{\textbf{Reducing class selectivity increases adversarial vulnerability in ResNet20 trained on CIFAR10.} (\textbf{a}) Test accuracy (y-axis) as a function of perturbation intensity ($\epsilon$; x-axis) and class selectivity regularization scale ($\alpha$; color) for the FGSM attack. (\textbf{b}) Test accuracy (y-axis) as a function of adversarial optimization steps (x-axis) and $\alpha$ for the PGD attack. (\textbf{c}) Network stability, as measured with input-output Jacobian (y-axis) as a function of $\alpha$.}
    \label{fig:si_adv_cifar10}
\end{figure*}

\begin{figure*}[!htbp]
    \centering
    \begin{subfloatrow*}
        \sidesubfloat[]{
        \label{fig:si_dim_per_unit_linear_cifar10_clean}
            \includegraphics[width=0.31\textwidth]{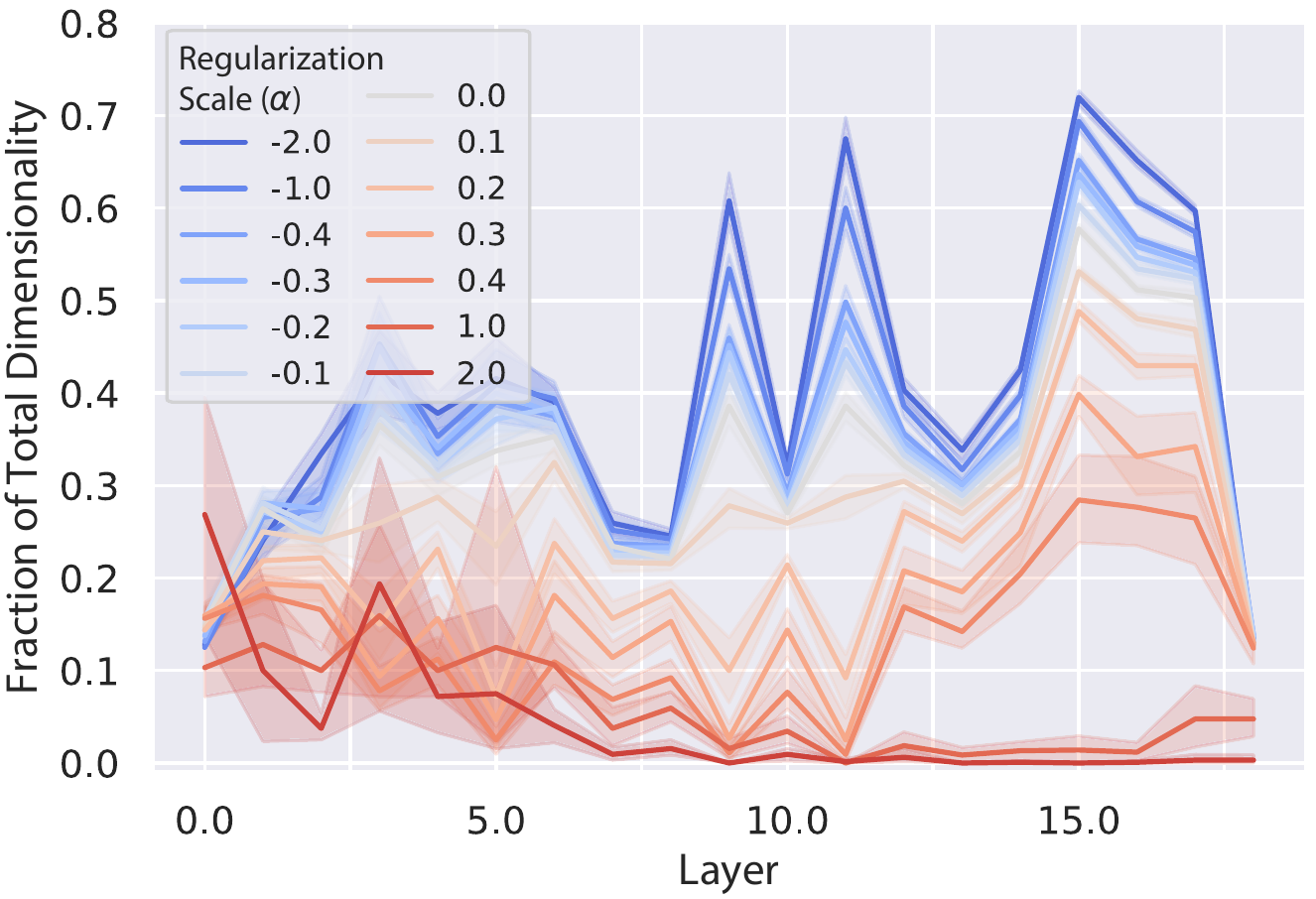}
        }
        \hspace{-6mm}
        \sidesubfloat[]{
            \label{fig:si_dim_per_unit_linear_cifar10c_diff}
            \includegraphics[width=0.31\textwidth]{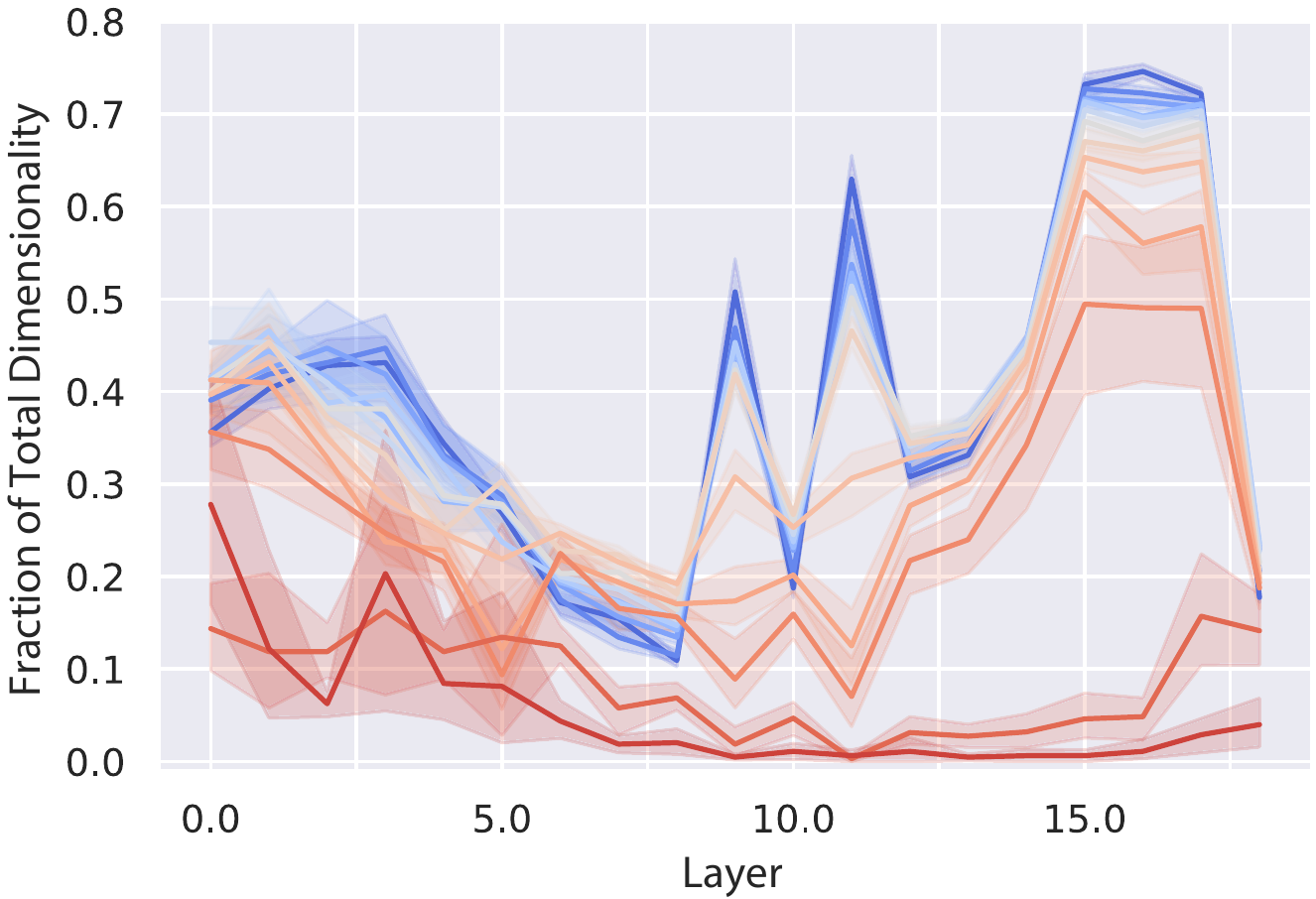}
        }        
        \hspace{1mm}
        \sidesubfloat[]{
            \label{fig:si_dim_per_unit_linear_cifar10_adv_diff}
            \includegraphics[width=0.31\textwidth]{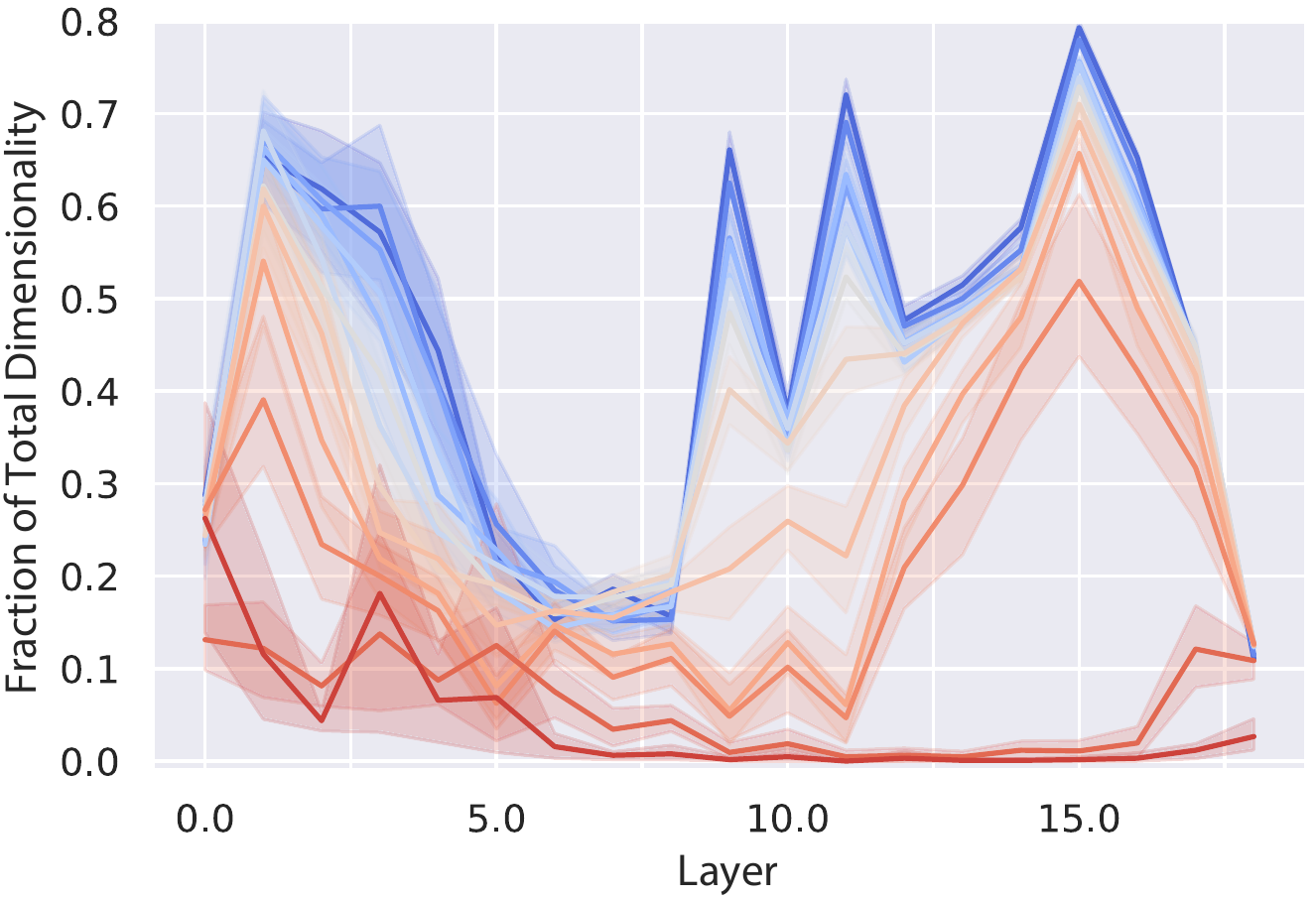}
        }
    \end{subfloatrow*}
    \vspace{-5mm}
    \caption{\textbf{Dimensionality in early layers predicts adversarial vulnerability in ResNet20 trained on CIFAR10.} (\textbf{a}) Fraction of dimensionality (y-axis; see Section \ref{sec:approach_dimensionality}) as a function of layer (x-axis). (\textbf{b}) Dimensionality of difference between clean and corrupted activations (y-axis) as a function of layer (x-axis). (\textbf{c}) Dimensionality of difference between clean and adversarial activations (y-axis) as a function of layer (x-axis).}
    \label{fig:si_dim_linear}
\end{figure*}



\end{document}